\pdfoutput=1

\documentclass[11pt]{article}

\usepackage[preprint]{acl}

\usepackage{times}
\usepackage{latexsym}

\usepackage[T1]{fontenc}

\usepackage[utf8]{inputenc}

\usepackage{microtype}

\usepackage{inconsolata}
\usepackage{graphicx} 
\usepackage{graphicx}
\usepackage[normalem]{ulem}
\useunder{\uline}{\ul}{}
\usepackage{booktabs}
\usepackage{multirow}
\usepackage{subfigure}
\usepackage{tcolorbox}
\newcommand{\ttbf}[1]{\texttt{#1}}
\newcommand{\name}[0]{\ttbf{Conifer}\space}

\title{Conifer: Improving Complex Constrained Instruction-Following Ability of Large Language Models}

\author{Haoran Sun\thanks{Equal contribution}, Lixin Liu\footnotemark[1], Junjie Li, Fengyu Wang, \\ {\bf Baohua Dong}, {\bf Ran Lin}, {\bf Ruohui Huang} \\
        Alibaba Group, Beijing, China \\
   \texttt{sunhaoran0402@gmail.com, llx271805@alibaba-inc.com}}

\begin{document}
\maketitle

\begin{abstract}
The ability of large language models (LLMs) to follow instructions is crucial to real-world applications. Despite recent advances, several studies have highlighted that LLMs struggle when faced with challenging instructions, especially those that include complex constraints, hindering their effectiveness in various tasks. To address this challenge, we introduce \textbf{Conifer}, a novel instruction tuning dataset, designed to enhance LLMs to follow multi-level instructions with complex constraints. Utilizing GPT-4, we curate the dataset by a series of LLM-driven refinement processes to ensure high quality. We also propose a progressive learning scheme that emphasizes an easy-to-hard progression, and learning from process feedback. Models trained with \name exhibit remarkable improvements in instruction-following abilities, especially for instructions with complex constraints. On several instruction-following benchmarks, our 7B model outperforms the state-of-the-art open-source 7B models, even exceeds the performance of models 10 times larger on certain metrics. All the code and \name dataset are available at \url{https://www.github.com/ConiferLM/Conifer}.
\end{abstract}

\section{Introduction}
\label{sec:intro}
Large language models (LLMs) have achieved impressive performance across a wide range of natural language processing (NLP) tasks. Instruction tuning, also known as supervised fine-tuning (SFT) \citep{ouyang_training_2022}, has enabled LLMs to better align with human preferences by following human instructions and generating helpful, honest, and harmless \citep{askell_general_2021} responses.

\begin{figure}
    \centering
    \includegraphics[width=0.9\linewidth]{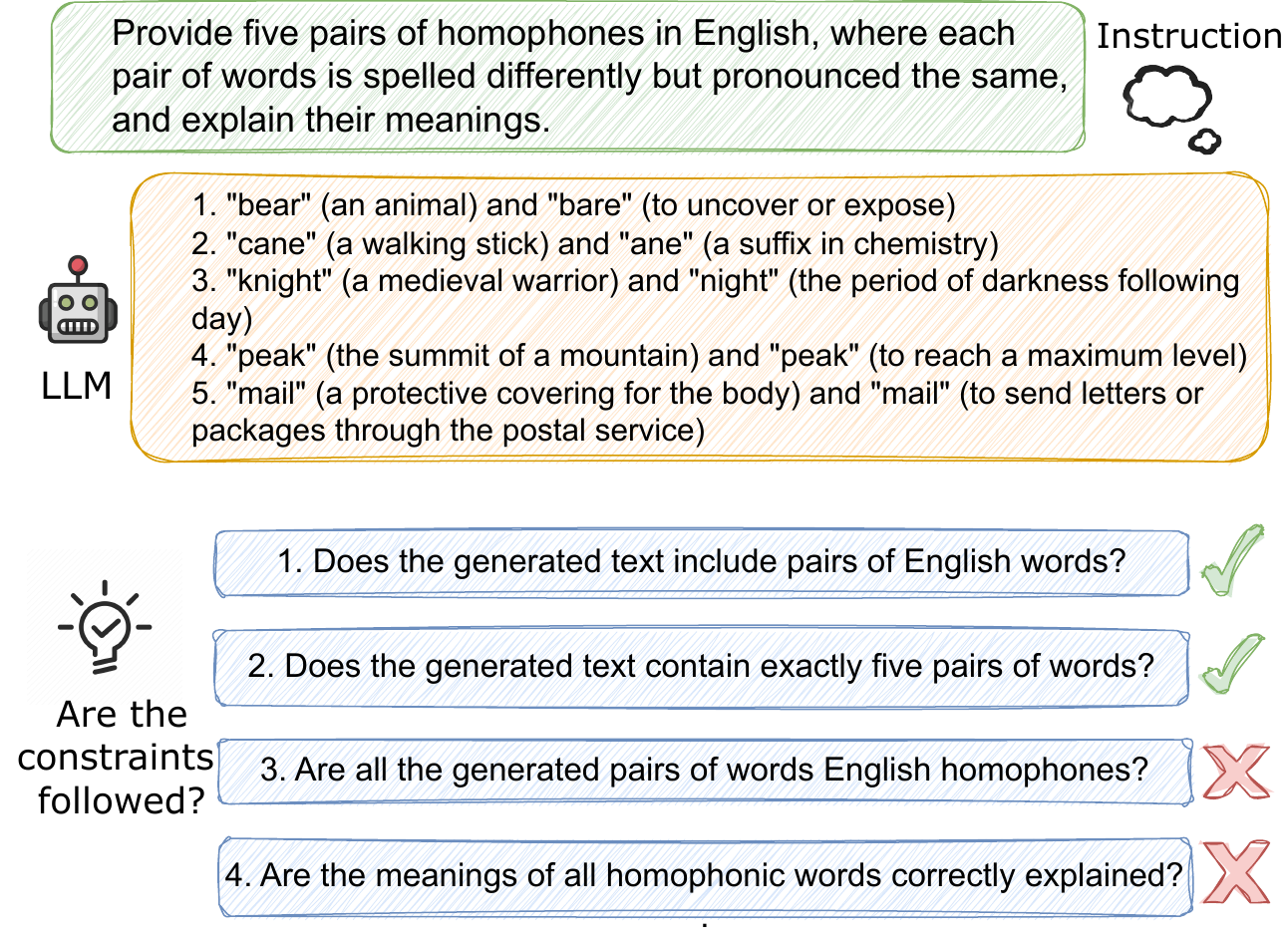}
    \caption{An example from InFoBench of how an LLM fails to follow the complex constraints in the instruction.}
    \label{example}
\end{figure}

Improving instruction-following abilities is a core direction of concern for the LLM community. Recent progress in the field has enabled open-source LLMs to demonstrate impressive performance in following instructions across numerous tasks \cite{li_alpacaeval_2023,zheng_judging_2023}. Nonetheless, LLMs, particularly those that are open source, still often struggle with more challenging tasks that include complex constraints in the instructions \cite{sun_evaluating_2023, jiang_followbench_2023, qin2024infobench}. Figure \ref{example} presents an instance illustrating an open source LLM's failure to adhere to instructions with multiple constraints. However, the challenge of enhancing LLMs to follow complex constraints is an area that remains insufficiently explored.

Recent studies have highlighted the importance of diverse and high-quality data in the success of instruction tuning \cite{zhou_lima_2023, liu2023makes}. Instead of relying on the costly process of manual data annotation, researchers are utilizing high-performing models like ChatGPT \cite{openai_chatgpt_2022} and GPT-4 \cite{openai_gpt-4_2023} to generate instruction tuning datasets \cite{taori_stanford_2023,chiang_vicuna_2023,peng_instruction_2023,xu_wizardlm_2023, ding_ultrachat_2023}. This approach demonstrates performance comparable to manually annotated high-quality data. Nonetheless, there is a lack of research in the area focusing on enhancing LLMs' capacity to handle complex constrained instructions. It is also challenging to prompt GPT-4 directly to generate instructions with complex constraints.

In this paper, we introduce a novel method for automatically generating an instruction-following dataset with complex constraints and construct the dataset called \includegraphics[width=0.15in]{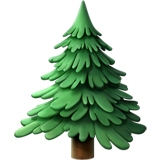} \textbf{\name} (Complex \textbf{Con}strained \textbf{I}nstruction \textbf{F}ollowing). This dataset is constructed using the capabilities of GPT-4 with a random selection of user queries as seeds from ShareGPT. To address the challenge of generating instructions with multiple complex constraints, we break the hard task into smaller, more manageable tasks. These include query reframing, constraint generation, recombination, and two-stage filtering processes. Additionally, to facilitate the instruction-tuned models to effectively learn the difficult tasks from data, we propose a progressive learning scheme. By organizing the data into a multi-turn conversational format following an ascending difficulty level, the model effectively learn through an easy-to-hard progression Moreover, the model is enabled to learn from both internal and external process feedback, utilizing GPT-4's insights on the explicit reasoning process required to follow complex constraints. 

The effectiveness of \name is validated by applying instruction tuning to the Mistral \cite{jiang_mistral_2023} and LLaMA-2 \cite{touvron_llama_2023-1} models, employing a combined dataset that merges 53k ShareGPT with 13k \name data, and further train the models with direct
preference optimization (DPO) \cite{rafailov_direct_2023}. The efficacy of our methodology is assessed on the recently proposed instruction-following benchmarks, including IFEval \cite{zhou_instruction-following_2023}, FollowBench \cite{jiang_followbench_2023} and InFoBench \cite{qin2024infobench}, where instructions are notably more complex and come with constraints. Additionally, it was benchmarked on widely-recognized AlpacaEval \cite{li_alpacaeval_2023} and MT-Bench \cite{zheng_judging_2023}. The  models trained with our \name dataset showed significant improvements in following complex and constrained instructions compared to its 7B counterparts. Notably, the performance of Conifer-7B-DPO even outperforms the best open-source models at the 70B scale on IFEval and FollowBench Hard. The main contributions of the paper are summarized as follows:
\begin{itemize}
    \item We tackle a critical yet under-explored challenge for LLMs: their difficulty in complying with complex, constrained instructions. 
    \item We introduce a new paradigm to generate the instructions with complex constraints, and release the \texttt{Conifer}, a novel instruction tuning dataset, specifically designed to enhance LLMs' capabilities in following complex constrained instructions. To construct this dataset, we decompose the complex task into smaller, more manageable tasks for GPT-4 to execute. We also introduce a progressive learning method that helps models to develop the ability to interpret instructions, by an easy-to-hard progression and learning from explicit internal and external process feedback.
    \item Extensive experiments, including ablation studies, demonstrate the effectiveness of the proposed approach. LLMs trained with the \name dataset achieve impressive performance in instruction-following ability, especially especially when dealing with complex and constrained instructions, as evidenced by various instruction-following benchmarks.
\end{itemize}

\begin{figure*}[t]
    \centering
    \includegraphics[width=1.0\linewidth]{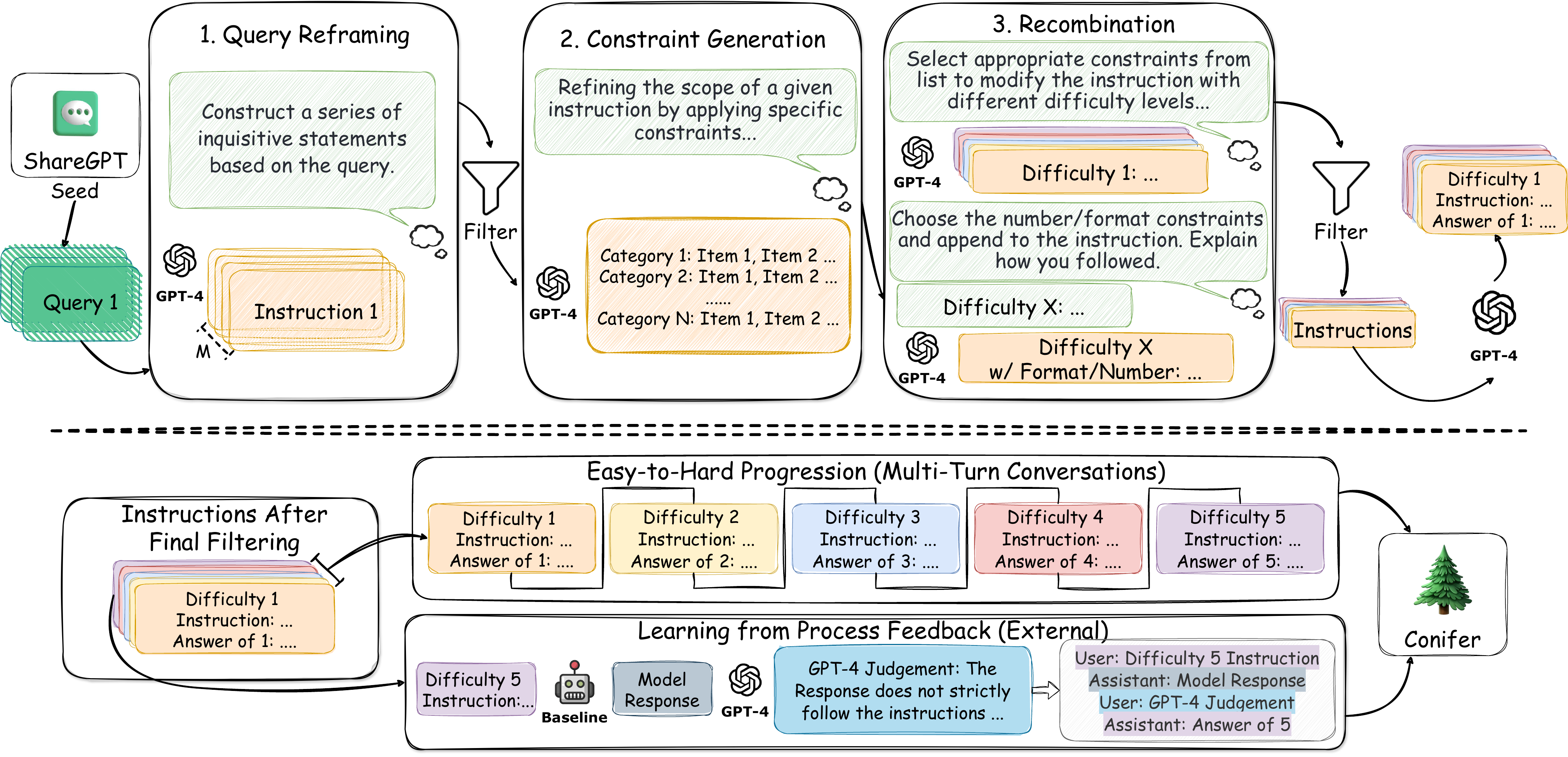}
    \caption{Paradigm of the production of the proposed Conifer dataset. The upper portion depicts the instruction collection phase (Section \ref{sec3.1}), while the lower portion outlines the progressive learning scheme (Section \ref{sec3.2}). }
    \label{Conifer}
\end{figure*}

\section{Related Work}
\paragraph{Instruction Tuning} Instruction tuning, or supervised fine-tuning (SFT) is to fine-tune LLMs to follow users' instructions, making them more controllable and predictable \cite{instruction_tuning_survey}. The SFT model can be further trained to better align with humans using reinforcement learning with human feedback (RLHF) \cite{ouyang_training_2022} or direct preference optimization (DPO) \cite{rafailov_direct_2023}.

Conventional approaches \citep{mishra_cross-task_2022, wei_finetuned_2021, sanh_multitask_2021, lou_muffin_2023} aggregate data from a large amount of NLP tasks to train multi-task models and that achieve strong results in downstream tasks. However, studies have highlighted a notable discrepancy between these NLP tasks and the actual way users interact with LLMs \citep{ouyang-etal-2023-shifted}, leading to responses less preferred by humans \citep{zheng_judging_2023, li_alpacaeval_2023}. In contrast, recent researches have been directed towards constructing instruction tuning datasets that better reflect humans' intent. These datasets can be either manually annotated \citep{ouyang_training_2022, kopf_openassistant_2023, zhou_lima_2023} or generated by advanced LLMs like GPT-4 \citep{peng_instruction_2023}. The generation of synthetic data through LLMs has emerged as a cost-effective research approach, highlighted by studies such as Alpaca \citep{taori_stanford_2023}, which utilizes self-instruct \citep{wang_self-instruct_2022} to prompt ChatGPT from a range of seed tasks. Vicuna \citep{chiang_vicuna_2023} and other studies \citep{geng_koala_2023, wang_openchat_2023} investigate the use of ShareGPT data, which more closely mirrors human intentions \citep{ouyang-etal-2023-shifted} and has demonstrated increased efficacy \citep{tulu2}. Additional methods leverage LLMs to simulate conversations \citep{ ding_enhancing_2023} or to undertake instruction expansion \cite{lou_muffin_2023} for generating SFT data. 

\paragraph{Instruction Following with Complex Instructions or Constraints} 
Traditional approaches for controllable text generation \cite{CTGSurvey2023} often involve fine-tuning LMs on specific tasks \cite{zhang_pointer_2020} or design post-processing strategies \cite{2019Plug}. In the era of LLM,  enhancing controllable text generation is often achieved by improving the instruction-following capabilities of LLMs. Several recent studies investigate the ability of LLMs for instruction following, presenting new benchmarks \citep{jiang_followbench_2023,qin2024infobench, zhou_instruction-following_2023}. These studies conclude that current LLMs struggle at meeting fine-grained or hard constraints \cite{sun_evaluating_2023}. Despite the significance of this challenge, few studies aim explicitly at addressing it. Most related, WizardLM \citep{xu_wizardlm_2023} aims to enhance the overall ability of LLMs with different task complexity. The proposed Evol-Instruct is created by in-breadth and in-depth transformation, which may alter the semantic of original instructions. But our paper mainly focuses on improving LLMs to follow instructions with complex constraints, achieved by adding constraints and preserving the core entities unchanged, so the scope of constraints here is substantially broader, more challenging, and more numerous than those in WizardLM.

\section{The \textbf{\name} Dataset}

In this section, we detail the methodologies employed in creating the Conifer dataset. We illustrate the complete data construction pipeline in Figure \ref{Conifer}, consisting of two principal stages: instruction collection from GPT-4 (Section \ref{sec3.1}) and organizing the dataset into a progressive learning scheme (Section \ref{sec3.2}). Throughout all stages involving the LLM, we utilize GPT-4 Turbo to execute the tasks.

\subsection{Instruction Collection} \label{sec3.1}
Building on the successful efforts of previous work \cite{taori_stanford_2023,xu_wizardlm_2023,lou_muffin_2023} and the analysis in Section \ref{sec:intro}, our goal is to generate instructions that are not only high in quality but also rich in diversity, complexity, and constraints. Therefore, we have randomly selected 6,000 user queries (prompts) from ShareGPT as our seed instructions. Given that ShareGPT data is sourced from open-domain conversations with ChatGPT, it offers a wide variety of topics, ensuring sufficient diversity of the Conifer dataset. Our preliminary experiments indicate that GPT-4 consistently struggles in generating instructions that contain multiple complex constraints. To address this challenge and to enhance the diversity and complexity of the Conifer dataset, we have decomposed this challenging task into smaller, more manageable tasks for GPT-4, including query reframing, constraint generation, recombination, and a two-stage filtering process.

\paragraph{Query Reframing} Our primary objective is to construct a dataset rich in complex constraints, targeting a diverse array of constraint types. To realize this, we initiate a process of query reframing to diversify the seed instructions. This involves editing the instructions to provide alternative perspectives while keeping the core entities unchanged. We utilize GPT-4 to reformulate each query into at least three distinct and varied forms, thereby enriching the dataset with a broader range of perspectives.

\paragraph{Constraint Generation} After the query reframing phase, we direct GPT-4 to identify the subjects and objects within the rephrased instructions and to generate a list of potential constraints to narrow down response options. Our preliminary tests have indicated that GPT-4 struggles to produce appropriate constraints consistently. Consequently, we have adopted in-context learning techniques, providing manually crafted examples that pair instructions with their corresponding constraints. These constraints are categorized into two levels: broader categories and specific items. For instance, in naming-related instructions, `Cultural Background' could serve as a category, with `Chinese', `Continental' as example items within this category. Organizing constraints in this way greatly assists us in prompting GPT-4 more effectively in the subsequent stages. 

\paragraph{Recombination} In this stage, for each instruction derived from the query reframing phase, GPT-4 is directed to select specific and suitable constraints from the generated constraint list and incorporate them into the instruction. This systematic modification facilitates the creation of instructions bound by predetermined constraints.

To increase the complexity of the instructions, we instruct GPT-4 to develop instructions with varying levels of difficulty. The level of difficulty is quantified by the number of constraints incorporated into each instruction. Each instruction of increasing difficulty should integrate 1-2 categories and 2-3 items, thus presenting a more complex challenge than its simpler counterpart. The scale of difficulty culminates at degree 5, indicating that each instruction derived from the query reframing stage will be crafted into up to five distinct instructions, reflecting a progressive scale of complexity.

Through observation, we notice that GPT-4 seems to struggle with incorporating format or numerical constraints with sentences while displaying a preference for adding content-based constraints. Since format and numerical constraints are important for instruction following, we randomly select 1,000 instructions after the recombination and prompt GPT-4 to augment them with additional format or numerical constraints. These constraints are synthesized by GPT-4 using three seed constraints. We curate multiple formatting and numerical constraints with the help of GPT-4. We manually review these synthesized constraints to ensure they are distinct from those in our evaluation benchmarks, after discarding any that are impractical or redundant, we obtained 32 constraints.

\paragraph{Two-Stage Filtering} To ensure the quality of the generated instructions, we incorporate a two-stage filtering process utilizing GPT-4. The initial filter occurs subsequent to the query reframing, primarily targeting the removal of instructions that lack necessary context for meaningful responses. The second filter takes place after the recombination, concentrating on identifying and resolving conflicts within instructions that may result from the recombination of various constraints.

By applying the outlined production process, we collect 35,613 instructions from GPT-4, and further engaged the model to generate corresponding responses for each instruction.

\subsection{Progressive Learning Scheme} \label{sec3.2}
Generating high-quality data with instructions and responses doesn't fully solve the problem of following complex instructions for LLMs. Given the dataset's high complexity, how to enable open-source, often smaller LLMs to learn how GPT-4 interpret complex instructions and tackle complex problems poses significant challenges. This is crucial ensure that these models don't merely imitating the style of proprietary language models \cite{gudibande2024false} but also their accuracy and depth of understanding. To tackle this challenge, we have developed a progressive learning strategy. This strategy uses two main techniques for structuring the data: an easy-to-hard progression and learning from process feedback.

\paragraph{Easy-to-Hard Progression} Inspired by curriculum learning \cite{bengio_curriculum_2009}, which suggests the benefits of starting with simpler examples and gradually moving to more complex ones, we have accordingly organized the Conifer dataset. Specifically, we aggregate multi-level instructions and answers generated from the same seed instruction into a single sample using the multi-turn conversational format. The sequence is organized to introduce the simpler tasks first, which then progressively lead towards more challenging ones, as depicted in Figure \ref{Conifer}.

\paragraph{Learning from Process Feedback} Preliminary experiments reveal that fine-tuned models struggle with adhering to the most challenging samples, particularly with respect to specific format and numerical constraints. Inspired by the recent research on process supervision \cite{lightman2023let}, which exposes the supervision signal at each reasoning steps and supervise the model's thought step, We facilitate the model to learn through process feedback from both internal and external perspectives. 

Internally, when prompted to respond to instructions that contain format and numerical constraints, GPT-4 is instructed to illustrate how it adheres to the specified constraints explicitly within its responses. The instructions of this part are randomly sampled from all the instructions of our Conifer. 

Externally, GPT-4 is tasked to identify constraints where it fails to adhere in the most challenging constraints, and this judgement is employed to construct a new multi-turn conversation, which take the form of: \{Difficult Instruction, Model Response, GPT-4 Judgement, GPT-4 Response\}. The instructions of this part are selectively sampled from the highest difficulty level, level 5, within the Conifer dataset.

By incorporating internal and external feedback by GPT-4's evaluations, models gain additional supervision signal on reasoning processes, enhancing their ability to interpret and follow complex constraints.

\subsection{Statistics of the Conifer Dataset}
Table \ref{tab:statistic} presents the Conifer dataset statistics, including the counts for the easy-to-hard progression ('Multi-Turn') and the learning from process feedback ('Feedback') parts. The 'Multi-Turn' part includes instructions with different degrees of difficulty (DD) ranging from 1 to 5 for the easy-to-hard progression, noting that not all multi-turn conversations contain all five levels. Additionally, the table lists learning from process feedback as internal and external feedback. There are totally 13,606 conversations in the Conifer dataset with an 3.02 average number of turns per conversation.

We also evaluate our Conifer's complexity and quality with the Deita complexity/quality scorer \cite{liu2023makes}. Relative to ShareGPT, Conifer demonstrates marked improvements in both aspects, as detailed in Figure \ref{fig:quality&complexity}.

Further analysis of the Conifer dataset, including examples of filtering cases, is shown in Appendix \ref{app:dataset}. The Conifer dataset has been released to the public on HuggingFace \url{https://huggingface.co/datasets/ConiferLM/Conifer}.

\begin{table}[t]
\centering
\footnotesize
\begin{tabular}{lclc}
\toprule
\multicolumn{1}{c}{Class}             & \# Samples       & \multicolumn{1}{c}{Subclass} & \# Ins. \\
\midrule
\multirow{5}{*}{Easy-to-Hard}               & \multirow{5}{*}{10302} & DD 1                 & 9167           \\
                                         &                       & DD 2                 & 8146           \\
                                         &                       & DD 3                 & 7157           \\
                                         &                       & DD 4                 & 6144           \\
                                         &                       & DD 5                 & 5017           \\
\midrule
\multirow{2}{*}{Process Feedback} & \multirow{2}{*}{3304} & Internal                     & 977            \\
                                         &                       & External                     & 2327          \\
\bottomrule
\end{tabular}
\caption{Statistics of the Conifer dataset. \# Ins. denotes the number of instructions.}
\label{tab:statistic}
\end{table}

\section{Experiments}
Comprehensive experiments are conducted to evaluate the performance of our proposed method and Conifer dataset, with a particular focus on improving the ability of models to follow complex constrained instructions.

\begin{table*}[t]
\centering
\resizebox{\textwidth}{!}{%
\begin{tabular}{lcc|c|cccccc|ccc}
\toprule
\multirow{2}{*}{Model} & \multirow{2}{*}{Base Model} & \multirow{2}{*}{Final Stage}  & \multicolumn{1}{c|}{IFEval}       & \multicolumn{6}{c|}{FollowBench (HSR)}                                                                                                                                               & \multicolumn{3}{c}{InFoBench}                                                 \\
\cmidrule(lr){5-10} \cmidrule(lr){11-13}
\multicolumn{1}{c}{}      & \multicolumn{1}{c}{}           & \multicolumn{1}{c}{}      & \multicolumn{1}{|c|}{loose prompt} & \multicolumn{1}{c}{Level 1} & \multicolumn{1}{c}{Level 2} & \multicolumn{1}{c}{Level 3} & \multicolumn{1}{c}{Level 4} & \multicolumn{1}{c}{Level 5} & \multicolumn{1}{c|}{Avg} & \multicolumn{1}{c}{Easy} & \multicolumn{1}{c}{Hard} & \multicolumn{1}{c}{Overall} \\
\midrule
GPT-4$^\dagger$                     & -                              & -                         & 79.3                             & 84.7                        & 76.1                        & 71.3                        & 74.5                        & 62.4                        & 73.8                    & 90.1                     & 89.1                     & 89.4                    \\
GPT-3.5 Turbo$^\dagger$                   & -                              & -                         & -                                & 80.3                        & 68.0                        & 68.6                        & 61.1                        & 53.2                        & 66.2                    & 90.4                     & 85.1                     & 86.7                    \\
\midrule
Qwen-72B-Chat$^\dagger$           & Qwen       & -   & 50.8                  & 73.8          & 63.3          & 54.3          & 45.2          & 39.9          & 55.3          & -             & -             & -             \\
LLaMA-2-70B-Chat$^\dagger$        & LLaMA-2    & RLHF   & -                     & 59.9          & 53.3          & 46.0          & 40.2          & 37.9          & 47.5          & 89.6          & 82.1          & 84.4          \\
\midrule
Vicuna-13B-v1.5          & LLaMA-2    & SFT   & 46.6                  & 71.2                     & 61.3                     & 48.3                     & 38.0                     & 33.1                     & 50.4                     & 85.7     & 73.7     & 77.3    \\
LLaMA-2-13B-ShareGPT     & LLaMA-2    & SFT   & 47.1                  & 59.2                     & 48.7                     & 45.8                     & 30.9                     & 27.4                     & 42.4                     & 84.3                     & 75.4                     & 78.2                     \\
Conifer-13B              & LLaMA-2    & SFT   & 47.5                  & 60.5                     & 53.6                     & 48.4                     & 40.7                     & 31.7                     & 47.0                     & 84.5     & 76.5     & 78.9    \\
\midrule

Deita-7B-V1.0-SFT        & Mistral    & SFT   & 45.1                  & 55.8          & 51.3          & 39.9          & 32.6          & 30.6          & 42.0          & 84.8          & 75.9          & 78.6          \\
Zephyr-7B-beta           & Mistral    & DPO   & 44.9                  & 57.6    & 51.9          & 41.9          & {\ul 41.4}   & 31.4          & 44.8          & 84.1          & 75.3          & 78.0          \\
Deita-7B-V1.0            & Mistral    & DPO   & {\ul 51.9}            & 55.8          & 49.3          & 46.7          & 39.6          & 37.3          & 45.7          & 86.2          & 78.6          & 80.9          \\
Mistral-7B-Muffin        & Mistral    & SFT   & 34.0                  & 40.1          & 31.5          & 23.9          & 17.8          & 14.4          & 25.6          & 70.0          & 66.9          & 67.8          \\
Mistral-7B-Evol-Instruct & Mistral    & SFT   & 44.0                  & 53.2          & 51.0          & 44.3          & 31.7          & 23.5          & 40.7          & 81.0          & 73.2          & 75.6          \\
Mistral-7B-ShareGPT      & Mistral    & SFT   & 43.4                  & 55.7          & 51.2          & 43.0          & 38.2          & 26.4          & 42.9          & 84.5          & 75.8          & 78.5          \\
Conifer-7B               & Mistral    & SFT   & 50.8                  & 54.3          & 49.5          & \textbf{49.3}          & 40.8          & 30.5          & 44.9          & 83.6          & 77.7          & 79.5          \\
Mistral-7B-ShareGPT-DPO  & Mistral    & DPO   & 48.2                  & {\ul 58.4}    & \textbf{53.9} & {\ul 48.3} & 39.1          & {\ul 38.6}    & {\ul 47.7}         & {\ul 86.8}          & {\ul 79.9}          & {\ul 82.0}          \\
Conifer-7B-DPO           & Mistral    & DPO   & \textbf{52.3}         & \textbf{60.3} & {\ul 53.6}    & 48.0       & \textbf{47.1} & \textbf{41.0} & \textbf{50.0} & \textbf{87.5} & \textbf{80.0} & \textbf{82.3}  \\
\bottomrule
\end{tabular}%
}
\caption{Main results on three instruction-following benchmarks: IFEval, FollowBench, and InFoBench. We use boldface for the best results and underline for the second-best results among the 7B models. $^\dagger$ indicates that the results are directly sourced from the original benchmarks.}
\label{tab:ins_fol_benchmark}
\end{table*}

\subsection{Models and Datasets}
We conduct experiments on two popular base LLMs, \textbf{Mistral-7B} \citep{jiang_mistral_2023} and \textbf{LLaMA-2-13B} \citep{touvron_llama_2023-1}. Mistral-7B is the state-of-the-art base large language model at the 7B parameter scale.

Since the Conifer dataset mainly address the instruction-following ability of LLMs under complex constraints, we merge the collected 13k Conifer dataset with the initial 53k ShareGPT\footnote{Collected from \url{https://huggingface.co/datasets/anon8231489123/ShareGPT_Vicuna_unfiltered} and after language filtering, we obtain 53k samples.} to form the final 66k SFT dataset, similar to the strategy of WizardLM \cite{xu_wizardlm_2023}. 

Although our primary focus is on the SFT alignment stage using the proposed Conifer dataset, we also explore the benefits of direct preference optimization (DPO) \cite{rafailov_direct_2023} training, which has shown significant improvements. Following previous work \cite{tunstall_zephyr_2023, liu2023makes}, we utilize the UltraFeedback dataset \cite{cui_ultrafeedback_2023} for DPO training.

\subsection{Evaluation}
\label{sec:evaluation}
We focused our evaluation on three challenging instruction-following benchmarks. Comparing with other benchmarks, instructions in these benchmarks are typically harder and often contains constraints.
\begin{itemize}
    \item \textbf{IFEval} \cite{zhou_instruction-following_2023} evaluates instruction-following LLMs with various verifiable instructions. Instructions from IFEval contain many lexical and format constraints. 
    
    \item \textbf{FollowBench} \cite{jiang_followbench_2023} is a multi-level fine-grained constraints following benchmark, including five different types (Content, Situation, Style, Format, and Example) and five difficulty levels, with performance assessed using GPT-4.

    \item \textbf{InFoBench} \cite{qin2024infobench} is a benchmark that breaks down a complex constraint into simpler criteria across multiple categories. It also leverages GPT-4 for evaluation process.
\end{itemize}

\begin{table*}[t]
\centering
\footnotesize
\begin{tabular}{lcc|cc|c}
\toprule
\multirow{2}{*}{Model}   & \multirow{2}{*}{Base Model} & \multirow{2}{*}{Stage} & \multicolumn{2}{c}{AlpacaEval 2.0} & \multicolumn{1}{c}{MT-Bench} \\
                         &                             &                        & LC Win Rate & Avg Length          &  Score   \\
\cmidrule(lr){1-6} 
GPT-4 0613$^\dagger$                    & -                           &                     & 30.2\% &   1140 & 9.18        \\
GPT-3.5 Turbo 0613$^\dagger$                  & -                           & -                  & 22.7\% &   1328 & 8.39         \\
\midrule
Deita-7B-v1.0-SFT$^\dagger$        & Mistral                     & SFT                    & -        &   - & 7.22            \\
Deita-7B-v1.0$^\dagger$            & Mistral                     & DPO     &   16.1\% &   1417   & \textbf{7.55}                  \\
Zephyr-7B-beta$^\dagger$           & Mistral                     & DPO            &  13.2\% &   1444  & 7.34              \\
\midrule
Mistral-7B-Muffin        & Mistral                     & SFT      &    3.8\%     &   736     & 3.90        \\
Mistral-7B-Evol-Instruct & Mistral                     & SFT     &   9.4\%    &   982      & 6.51          \\
Mistral-7B-ShareGPT      & Mistral                     & SFT     &    11.6\%   &   1070      & 6.86    \\
Conifer-7B               & Mistral                     & SFT     &     12.5\%   &   1052   & 7.08              \\
Mistral-7B-ShareGPT-DPO  & Mistral                     & DPO     &   15.1\%   &   1276    & 7.10    \\
Conifer-7B-DPO           & Mistral                     & DPO    &     \textbf{17.1\%}  &   1253   & 7.25        \\
\bottomrule
\end{tabular}
\caption{Evaluation on the AlpacaEval and MT-Bench benchmarks for general LLM instruction-following ability. $^\dagger$ indicates that the scores are directly sourced from the original benchmarks.}
\label{tab:dialo_benchmark}
\end{table*}

We also conduct evaluations using two widely recognized instruction-following benchmarks on LLMs' general ability to align with human preferences. 
\begin{itemize}
    \item \textbf{AlpacaEval} \cite{li_alpacaeval_2023} is an LLM-based automatic evaluation by GPT-4 Turbo. The score is the win-rate against the reference model (GPT-4 Turbo) on the dataset. We adopt AlpacaEval 2.0 Length-Adjusted (LC) win rate \cite{alpaca_eval_length}, the latest version of AlpacaEval, which has a spearman correlation of 0.98 with ChatBot Arena.
    \item \textbf{MT-Bench} \cite{zheng_judging_2023} is multi-turn question set containing 80 multi-turn questions. The model answers an initial question, and then responds to a predefined another question. The responses are rated by GPT-4 on scores scaled from 1-10. We follow the guidelines from the authors to use GPT-4 0314 version for judgement.
\end{itemize}
Alignment with human preference is the main objective for instruction tuning. The two benchmarks are effective measures for evaluating how well models align with general user preferences, showing high correlation with human judgments.

We compare our Conifer models with various open-source LLMs. Specifically, we choose two 70B size models: Qwen-72B-Chat \cite{qwen} and LLaMA-2-70B-Chat \cite{touvron_llama_2023-1} which are among the best open-source LLMs. All the compared open-source 7B LLMs have released their instruction tuning data, including Deita \cite{liu2023makes} and Zephyr-7B-beta \cite{tunstall_zephyr_2023} models which establish SOTA results through DPO alignment based on Mistral-7B. In 13B experiments, we compare with Vicuna-13B-v1.5 \cite{chiang_vicuna_2023} and our trained LLaMA-2-13B on 53k ShareGPT data only, since Vicuna's 125k ShareGPT dataset  has not been released yet.

Conifer is also compared with two researches that are most related to ours: WizardLM \cite{xu_wizardlm_2023}, and Muffin \cite{lou_muffin_2023}. We train baseline models using only the ShareGPT dataset, the Muffin dataset, and as well as combining ShareGPT with Evol-Instruct (following the WizardLM paper), keeping the base LLM and experimental settings consistent with ours for fair comparisons.

Other experimental details, including (1) details of training, (2) details of baselines, (3) full results on the IFEval and FollowBench, and (4) additional results on the Open LLM leaderboard are shown in Appendix \ref{app:exp}.

\subsection{Main Results}
Table \ref{tab:ins_fol_benchmark} presents our results on three instruction-following benchmarks: IFEval, FollowBench, and InFoBench. The Conifer-7B-DPO model exhibits exceptional performance across all benchmarks, even outperforming or matching the capabilities of 70B models. Specifically, it achieves state-of-the-art performance on the IFEval benchmark, and achieves an excellent average score on FollowBench, surpassing LLaMA-2-70B-Chat and trailing just behind Qwen-72B-Chat. Our model's proficiency is particularly evident in handling difficult constraints, with its excellent performance in the Level 4 to Level 5 constraint difficulty on FollowBench. Specifically, Conifer-7B-DPO obtains a 41.0\% success rate on Level 5, outperforming Qwen-72B-Chat's 39.9\%. On InFoBench, it continues its remarkable performance against all open-source models of similar size, and close the gap between 7B models and LLaMA-2-70B-Chat, confirming Conifer's superiority in handling challenging tasks.

Within the 7B model category, our primary comparisons involve state-of-the-art Mistral-7B models like Zephyr and Deita, as well as WizardLM (Evol-Instruct) and Muffin. Comparisons reveal that Conifer-7B outperforms all the aforementioned models in a substantial margin. Compared to Deita-7B and Mistral-7B-Evol-Instruct, which are trained on datasets noted for their high quality and complexity, our Conifer-7B showcases remarkable superiority, affirming the exceptional performance of the Conifer dataset in improving the ability of LLMs in follow constrained instruction-following tasks.

Comparing to models trained with ShareGPT dataset only, Conifer surpasses them across all benchmarks with the same 7B or 13B base LLM, under both the SFT and DPO stages. Note that models we trained on LLaMA-2-13B do not surpass Vicuna-13B-v1.5 on FollowBench, which may be attributed to Vicuna's use of more ShareGPT data (125k) that has not been made public.

In Table \ref{tab:dialo_benchmark}, we also illustrate the performance on AlpacaEval and MT-Bench benchmarks that mainly assess LLMs’ overall ability to align with human preferences. From this table, our Conifer model outperforms baseline models at the same stage, such as ShareGPT, Evol-Instruct, and Muffin, indicating consistent improvements in aligning with general human preferences. 

While AlpacaEval and MT-Bench benchmarks may correlate with response lengths \cite{zhao2024long}, Conifer models do not generate lengthy responses. The average length is much shorter than those produced by Deita and Zephyr models. The latest AlpacaEval LC win rate solves the length bias issue. Conifer-7B-DPO achieves a high 17.1\% LC Win Rate, outperforming other open-source 7B models like Zephyr-7B-beta and Deita-7B-v1.0. These validate Conifer's competitive performance within the scope of open-source 7B models.

\begin{table}[t]
\centering
\footnotesize
\begin{tabular}{l|c|ccc}
\toprule
\multirow{2}{*}{Model} & \multirow{2}{*}{IFEval} & \multicolumn{3}{c}{FollowBench}         \\
\cmidrule(lr){3-5}
                       &                         & L1-L3 & L4-L5 & \multicolumn{1}{l}{Avg} \\
\midrule
w/o Conifer                                & 43.4                                        & 50.0      & 32.3      & 42.9    \\
\midrule
random shuffle                             & 48.6                                        & 48.1      & 38.2      & 44.1    \\
hard-to-easy                               & 49.9                                        & 47.5      & 33.1      & 41.7    \\
easy only                                  & 47.5                                        & 51.3      & 33.5      & 44.2    \\
hard only                                  & 48.2                                        & 49.5      & 34.3      & 43.4    \\
single turn                                & 46.2                                        & 51.5      & 33.1      & 44.2    \\
\midrule
w/o process feedback                       & 44.2                                        & 48.4      & 36.1      & 43.5    \\
w/ internal only                           & 49.4                                        & 48.0      & 37.2      & 43.7    \\
w/ external only                           & 48.2                                        & 51.6      & 35.7      & 45.3    \\
w/o format\&number                         & 48.6                                        & 51.4      & 35.2      & 44.9    \\
\midrule
Conifer                                    & 50.8                                        & 51.0      & 35.6      & 44.9    \\
\bottomrule
\end{tabular}
\caption{Ablation studies results when applying different progressive learning scheme in Section \ref{sec3.2}.}
\label{tab:ablation_study}
\end{table}

\subsection{Ablation Studies}
Ablation studies are carried out to investigate the impact of the progressive learning scheme introduced in Section \ref{sec3.2}, with Mistral-7B as base LLM and applying SFT for alignment. 

The results, summarized in Table \ref{tab:ablation_study}, 
reveal that varying the sequence of difficulty within the easy-to-hard progression, such as through random shuffling or reversing the order to hard-to-easy, results in reduced overall performance. In particular, randomizing the sequence results in a notable decrease in performance. Excluding certain difficulty levels, such as including only easy or only hard instructions, leads to decreased performance on the L4-L5 or L1-L3 of FollowBench, respectively. Removing the easy-to-hard progression learning scheme, by using a single turn instead of a multi-turn format, diminishes the model's effectiveness on instruction following capability.

The significance of learning from process feedback is also completely examined. It is clear that each component plays a crucial role, as demonstrated in Table \ref{tab:ablation_study}. Note that relying solely on internal feedback results in a notable decrease in FollowBench, while the IFEval score slightly improves, which is in line with expectations since internal feedback only addresses format and numerical constraints where IFEval is particularly focused. Similarly, the slight performance boost at the L4-L5 level of FollowBench when only external feedback is provided can align with the understanding that such feedback is most effective for the hardest instructions. Incorporating both internal and external process feedback is able to achieve comprehensive performance improvements across IFEval and FollowBench.

\begin{table}[t]
\centering
\footnotesize
\begin{tabular}{llcc}
\toprule
Train Set & Test Set          & \begin{tabular}[c]{@{}c@{}}Rephrased\\ Samples\end{tabular} & \begin{tabular}[c]{@{}c@{}}Percentage\\ (\%)\end{tabular} \\
\midrule
Conifer   & IFEval      & 3                 & 0.55                        \\
Conifer   & FollowBench & 5                 & 0.53                        \\
Conifer   & InFoBench   & 11                & 2.20                        \\
\midrule
ShareGPT  & IFEval      & 29                & 5.36                        \\
ShareGPT  & FollowBench & 59                & 6.25                        \\
ShareGPT  & InFoBench   & 36                & 7.20                        \\
\bottomrule
\end{tabular}
\caption{Rephrased samples assessment between Conifer and ShareGPT datasets with the benchmarks, as evaluated by GPT-4 Turbo. The values indicate the proportion of rephrased samples within each test set.}
\label{tab:rephrase}
\end{table}

\begin{figure}[t]
    \centering
    \includegraphics[width=1.0\linewidth]{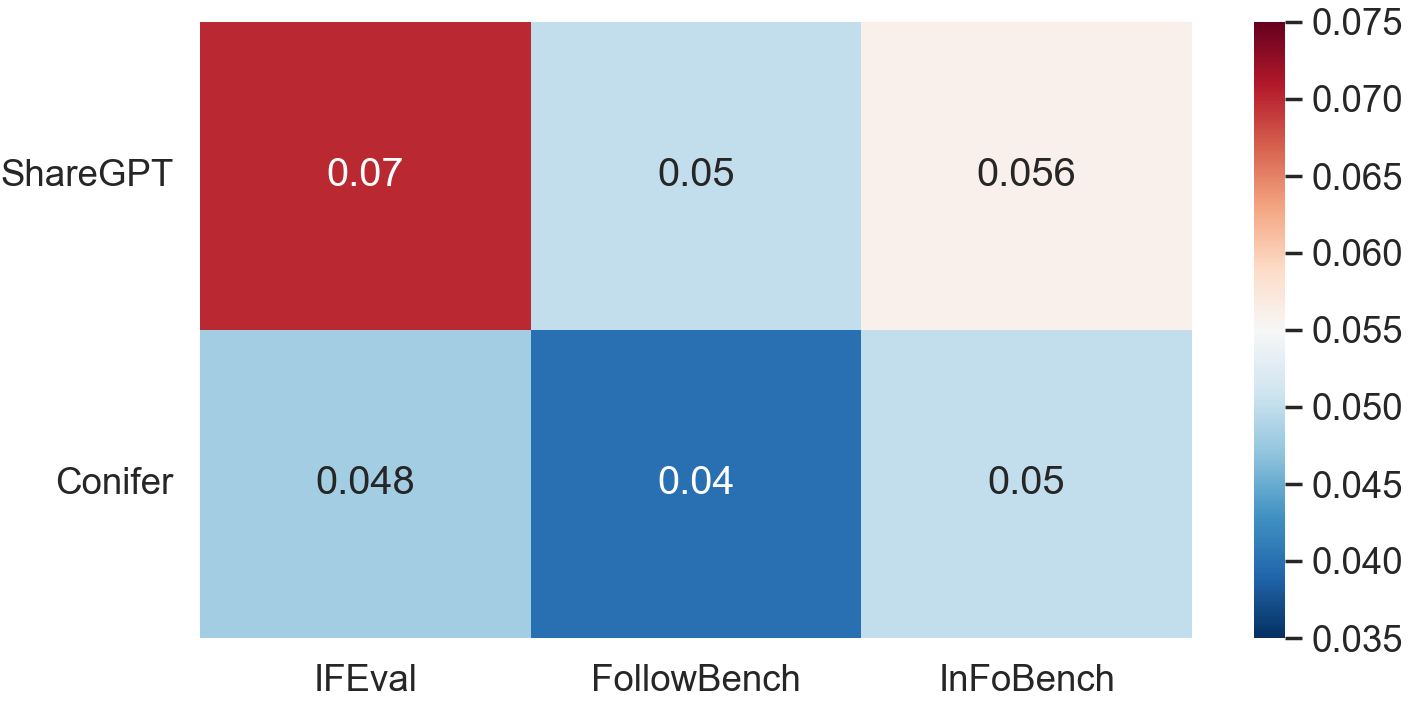}
    \caption{Cosine similarities between sentences from training and testing sets. Values near 0 indicate little overlap and no data contamination.}
    \label{cos_sim}
\end{figure}

\begin{figure*}[t] 
	\centering
	\subfigure[Distribution of Turn]{
	    \centering
		\label{fig:round}
		\includegraphics[width=0.3\linewidth]{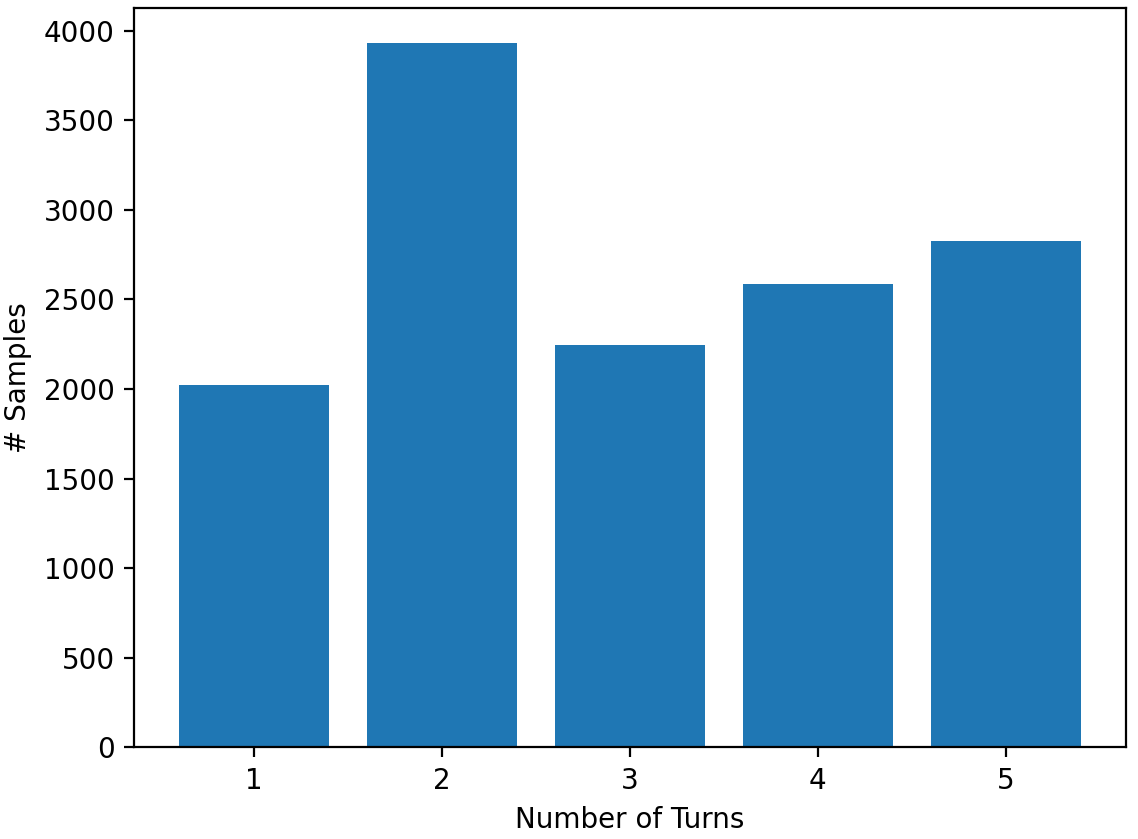}
        }\quad
	\subfigure[Distribution of Instruction Length]{
	    \centering
		\label{fig:ins}
		\includegraphics[width=0.3\linewidth]{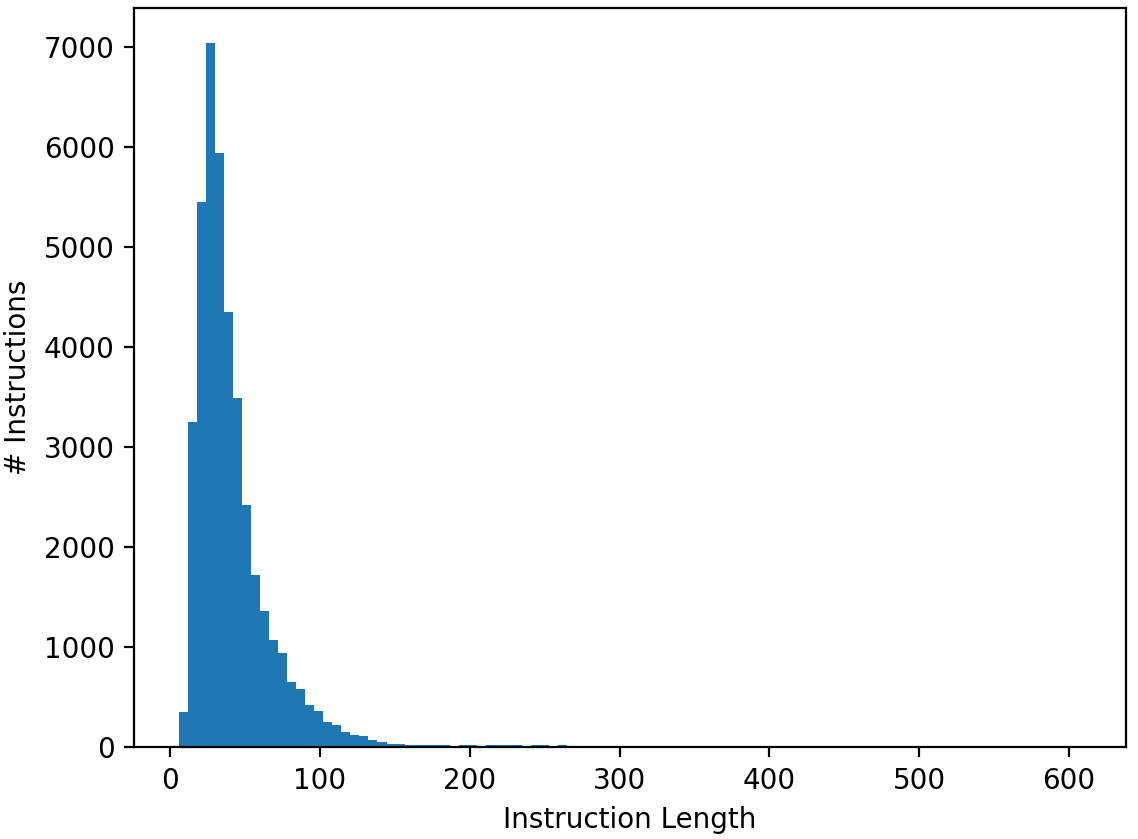}
        }\quad
	\subfigure[Distribution of Response Length]{
	    \centering
		\label{fig:resp}
		\includegraphics[width=0.3\linewidth]{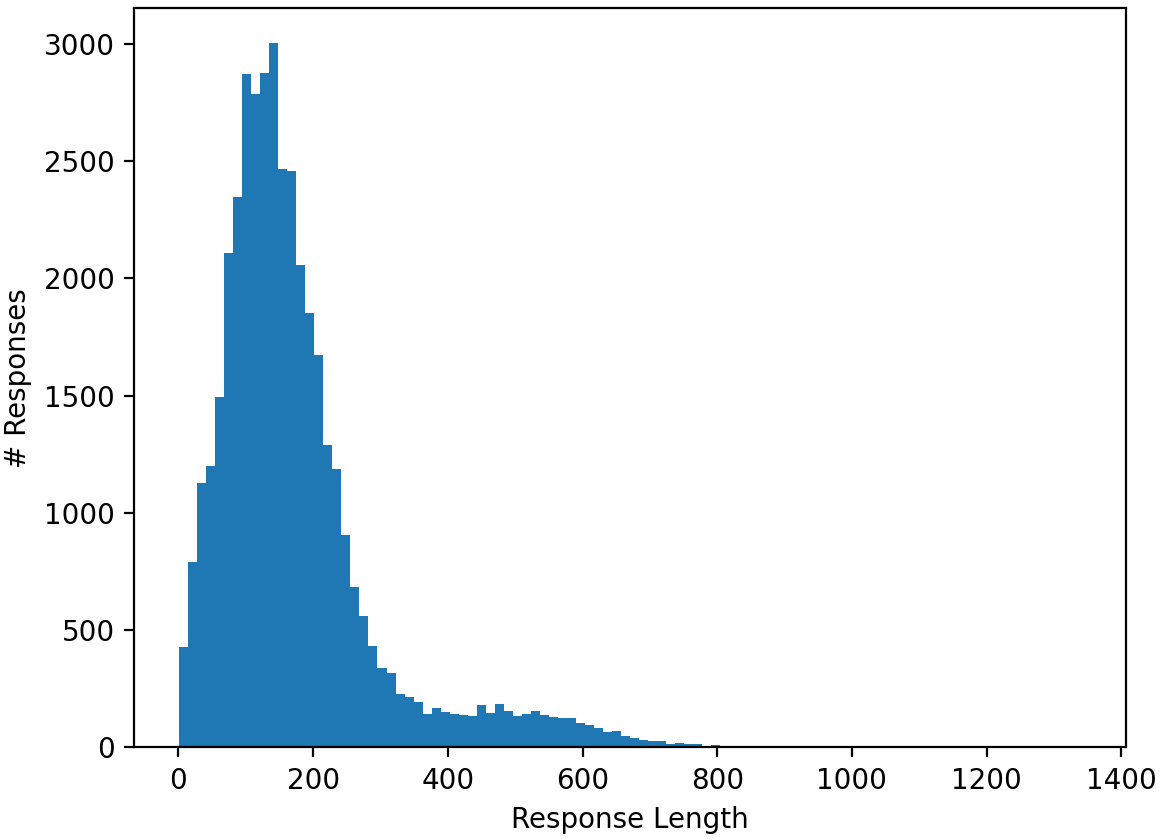}
        }
	\caption{Distribution of turns round, instructions length, and responses length in Conifer dataset.}
	\label{fig:quantity}
\end{figure*}

\begin{figure*}[t] 
	\centering
	\subfigure[Conifer Complexity Score]{
	    \centering
		\label{fig:conifer_complexity}
		\includegraphics[width=0.35\linewidth]{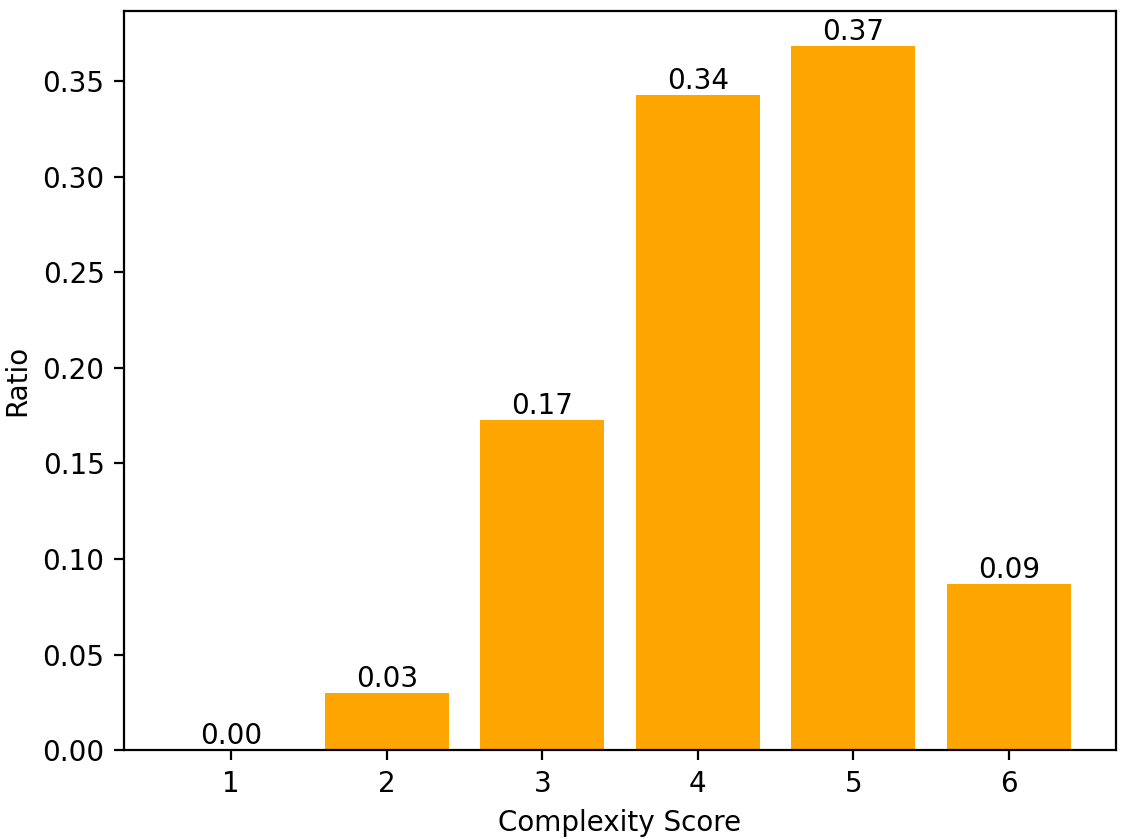}
        }\quad
	\subfigure[ShareGPT Complexity Score]{
	    \centering
		\label{fig:sharegpt_complexity}
		\includegraphics[width=0.35\linewidth]{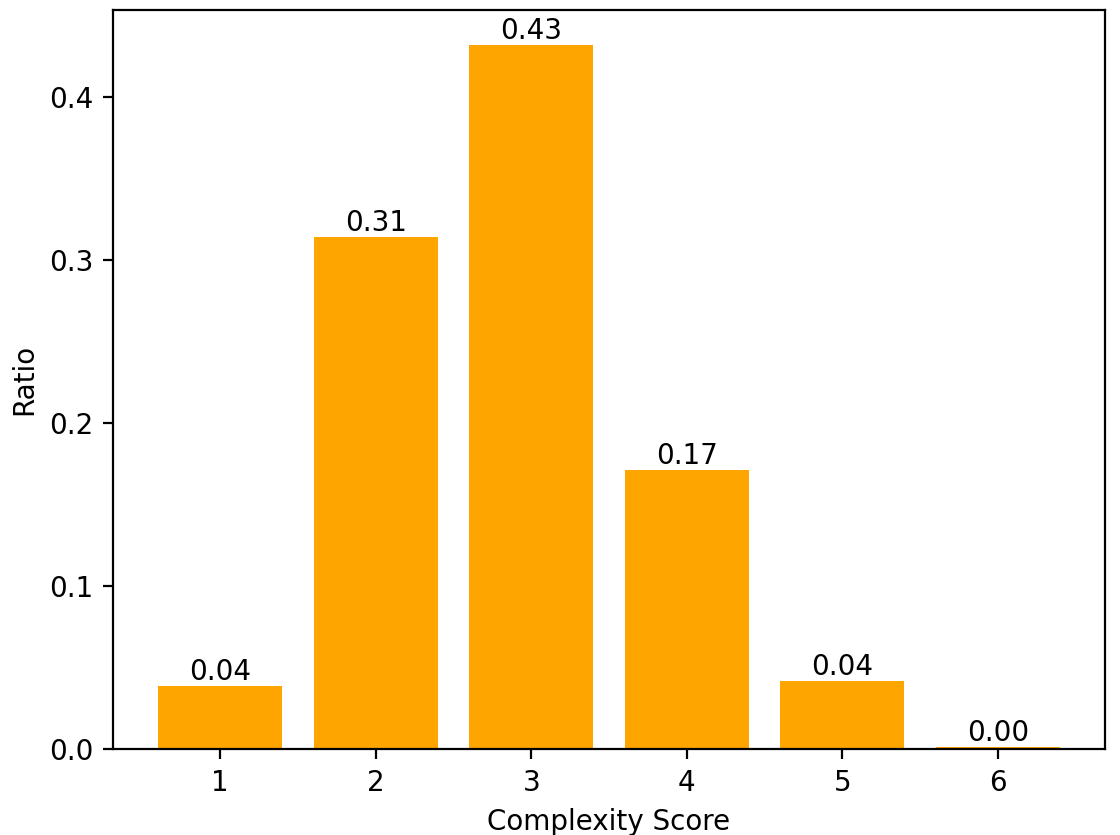}
        }\\
	\subfigure[Conifer Quality Score]{
	    \centering
		\label{fig:conifer_quality}
		\includegraphics[width=0.35\linewidth]{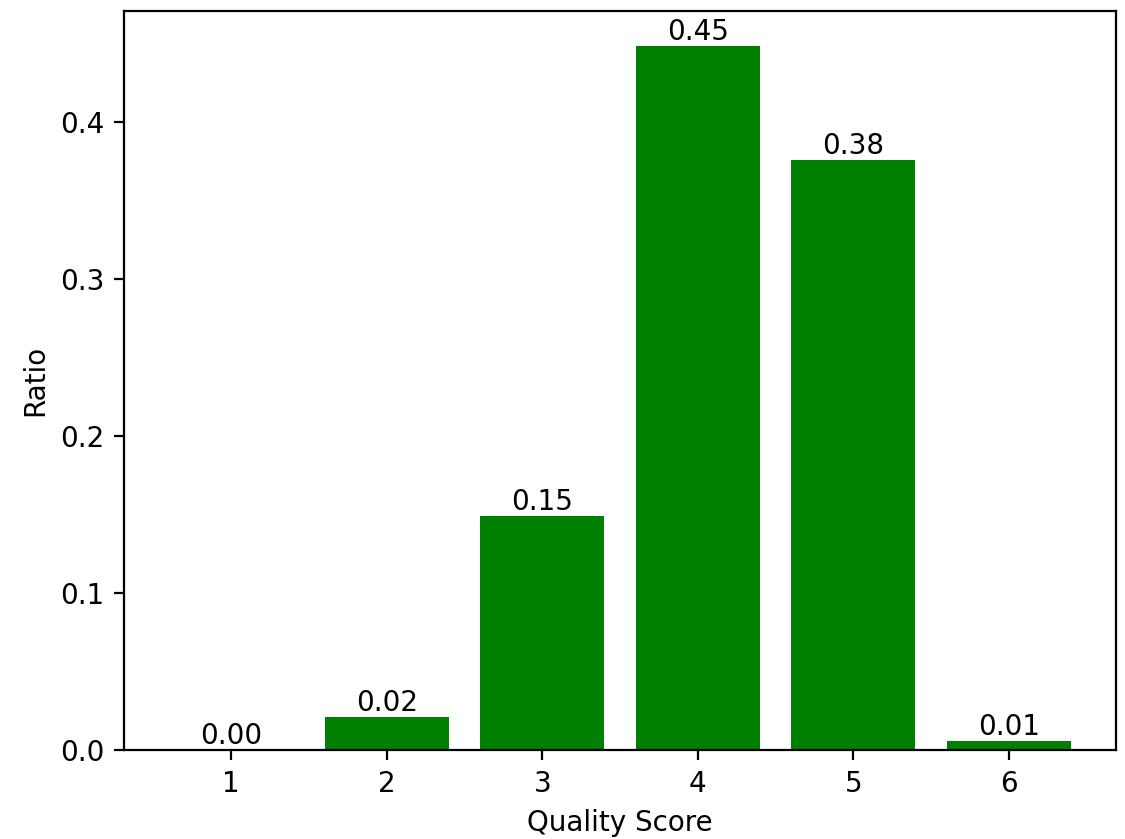}
        }\quad
	\subfigure[ShareGPT Quality Score]{
	    \centering
		\label{fig:sharegpt_quality}
		\includegraphics[width=0.35\linewidth]{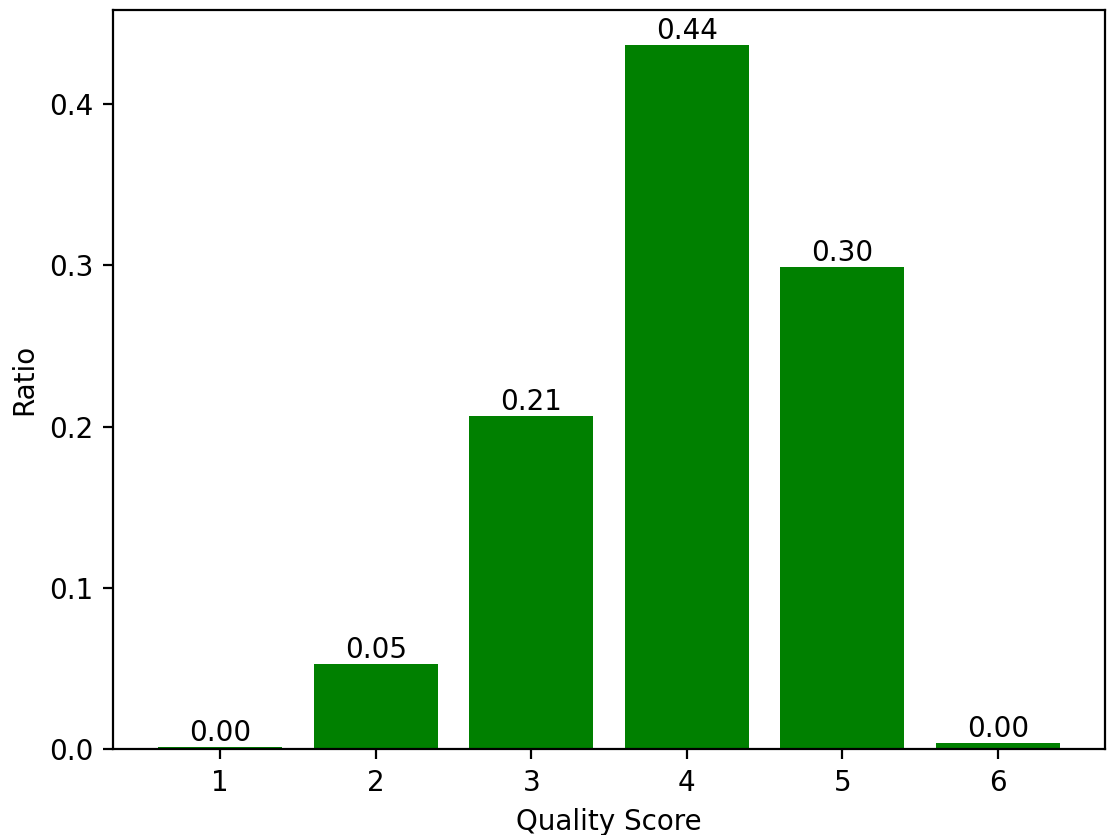}
        }
	\caption{Distribution of complexity/quality score of our Conifer dataset and ShareGPT data, which is measured by Deita complexity/quality scorer.}
	\label{fig:quality&complexity}
\end{figure*}

\subsection{Is There Any Data Contamination?}
Data contamination has been a critical problem in the era of LLMs, where training on the test set will definitely bring improvement \cite{wei2023skywork,deng2023investigating}. To assess whether Conifer dataset exhibits data contamination with three instruction-following benchmarks, we conduct analysis using: (1) cosine similarity between Conifer and test samples \cite{lou_muffin_2023}; (2) using GPT-4 to detect rephrased samples across training and testing sets \cite{yang2023rethinking}.

Figure \ref{cos_sim} illustrates sentence-level cosine similarities of embeddings, which is calculated by Sentence Transformers\footnote{\url{https://huggingface.co/sentence-transformers/all-MiniLM-L6-v2}}.
We compare the Conifer against ShareGPT across the IFEval, FollowBench, and InFoBench test sets and get lower similarity scores, suggesting minimal data overlap. 

Additionally, following recent work on data contamination detection \cite{yang2023rethinking}, we employ GPT-4 to identify rephrased samples from Conifer dataset and the benchmarks, results are in Table \ref{tab:rephrase}. Compared to ShareGPT dataset, Conifer shows relatively lower percentage of similar samples, which indicates an absence of data contamination.

\subsection{Quantity, Quality, and Complexity}
To give a better illustration of the quantity of our Conifer dataset, the distribution of the number of turns, instruction length, and response length in our Conifer, is shown in Figure \ref{fig:quantity}. All the length is counted by words using NLTK toolkit.

To examine the quality and complexity of our Conifer, we utilize the Deita quality/complexity scorer \cite{liu2023makes} which is trained based on LLaMA. The distribution of quality and complexity are shown in Figure \ref{fig:quality&complexity}. It should be noted that the scorer might have bias. We also calculate both scores of ShareGPT data. From the comparison, we can conclude that Conifer exhibit more complexity and higher quality than ShareGPT.

\section{Conclusion}
In this paper, we tackle the important yet under-explored challenge for LLMs: difficulty in following complex, constrained instructions. We create a novel instruction tuning dataset, Conifer, with GPT-4's assistance, and carefully curate it to ensure quality.  We propose an effective progressive learning scheme, with easy-to-hard progression and learning from progress feedback. Experimental results demonstrate that
Conifer-7B-DPO outperforms the best open-source 7B models according to instruction-following benchmarks, even matches the performance of models 10 times larger on certain metrics.. 

\bibliography{main}

\begin{thebibliography}{54}
\expandafter\ifx\csname natexlab\endcsname\relax\def\natexlab#1{#1}\fi

\bibitem[{ai2(2019)}]{ai2:winogrande}
 2019.
\newblock Winogrande: An adversarial winograd schema challenge at scale.

\bibitem[{Askell et~al.(2021)Askell, Bai, Chen, Drain, Ganguli, Henighan, Jones, Joseph, Mann, DasSarma, and {others}}]{askell_general_2021}
Amanda Askell, Yuntao Bai, Anna Chen, Dawn Drain, Deep Ganguli, Tom Henighan, Andy Jones, Nicholas Joseph, Ben Mann, Nova DasSarma, and {others}. 2021.
\newblock A general language assistant as a laboratory for alignment.
\newblock \emph{arXiv preprint arXiv:2112.00861}.

\bibitem[{Bai et~al.(2023)Bai, Bai, Chu, Cui, Dang, Deng, Fan, Ge, Han, Huang, Hui, Ji, Li, Lin, Lin, Liu, Liu, Lu, Lu, Ma, Men, Ren, Ren, Tan, Tan, Tu, Wang, Wang, Wang, Wu, Xu, Xu, Yang, Yang, Yang, Yang, Yao, Yu, Yuan, Yuan, Zhang, Zhang, Zhang, Zhang, Zhou, Zhou, Zhou, and Zhu}]{qwen}
Jinze Bai, Shuai Bai, Yunfei Chu, Zeyu Cui, Kai Dang, Xiaodong Deng, Yang Fan, Wenbin Ge, Yu~Han, Fei Huang, Binyuan Hui, Luo Ji, Mei Li, Junyang Lin, Runji Lin, Dayiheng Liu, Gao Liu, Chengqiang Lu, Keming Lu, Jianxin Ma, Rui Men, Xingzhang Ren, Xuancheng Ren, Chuanqi Tan, Sinan Tan, Jianhong Tu, Peng Wang, Shijie Wang, Wei Wang, Shengguang Wu, Benfeng Xu, Jin Xu, An~Yang, Hao Yang, Jian Yang, Shusheng Yang, Yang Yao, Bowen Yu, Hongyi Yuan, Zheng Yuan, Jianwei Zhang, Xingxuan Zhang, Yichang Zhang, Zhenru Zhang, Chang Zhou, Jingren Zhou, Xiaohuan Zhou, and Tianhang Zhu. 2023.
\newblock Qwen technical report.
\newblock \emph{arXiv preprint arXiv:2309.16609}.

\bibitem[{Bengio et~al.(2009)Bengio, Louradour, Collobert, and Weston}]{bengio_curriculum_2009}
Yoshua Bengio, Jérôme Louradour, Ronan Collobert, and Jason Weston. 2009.
\newblock Curriculum learning.
\newblock In \emph{Proceedings of the 26th annual international conference on machine learning}, pages 41--48. ACM.

\bibitem[{Chiang et~al.(2023)Chiang, Li, Lin, Sheng, Wu, Zhang, Zheng, Zhuang, Zhuang, Gonzalez, Stoica, and Xing}]{chiang_vicuna_2023}
Wei-Lin Chiang, Zhuohan Li, Zi~Lin, Ying Sheng, Zhanghao Wu, Hao Zhang, Lianmin Zheng, Siyuan Zhuang, Yonghao Zhuang, Joseph~E. Gonzalez, Ion Stoica, and Eric~P. Xing. 2023.
\newblock \href {https://lmsys.org/blog/2023-03-30-vicuna/} {Vicuna: {An} {Open}-{Source} {Chatbot} {Impressing} {GPT}-4 with 90\%* {ChatGPT} {Quality}}.
\newblock Published: Blog post.

\bibitem[{Clark et~al.(2018)Clark, Cowhey, Etzioni, Khot, Sabharwal, Schoenick, and Tafjord}]{clark_think_2018}
Peter Clark, Isaac Cowhey, Oren Etzioni, Tushar Khot, Ashish Sabharwal, Carissa Schoenick, and Oyvind Tafjord. 2018.
\newblock Think you have {Solved} {Question} {Answering}? {Try} {ARC}, the {AI2} {Reasoning} {Challenge}.
\newblock \emph{arXiv:1803.05457v1}.

\bibitem[{Cobbe et~al.(2021)Cobbe, Kosaraju, Bavarian, Chen, Jun, Kaiser, Plappert, Tworek, Hilton, Nakano, Hesse, and Schulman}]{cobbe2021gsm8k}
Karl Cobbe, Vineet Kosaraju, Mohammad Bavarian, Mark Chen, Heewoo Jun, Lukasz Kaiser, Matthias Plappert, Jerry Tworek, Jacob Hilton, Reiichiro Nakano, Christopher Hesse, and John Schulman. 2021.
\newblock Training verifiers to solve math word problems.
\newblock \emph{arXiv preprint arXiv:2110.14168}.

\bibitem[{Cui et~al.(2023)Cui, Yuan, Ding, Yao, Zhu, Ni, Xie, Liu, and Sun}]{cui_ultrafeedback_2023}
Ganqu Cui, Lifan Yuan, Ning Ding, Guanming Yao, Wei Zhu, Yuan Ni, Guotong Xie, Zhiyuan Liu, and Maosong Sun. 2023.
\newblock Ultrafeedback: {Boosting} language models with high-quality feedback.
\newblock \emph{arXiv preprint arXiv:2310.01377}.

\bibitem[{Dathathri et~al.(2020)Dathathri, Madotto, Lan, Hung, Frank, Molino, Yosinski, and Liu}]{2019Plug}
Sumanth Dathathri, Andrea Madotto, Janice Lan, Jane Hung, Eric Frank, Piero Molino, Jason Yosinski, and Rosanne Liu. 2020.
\newblock \href {https://openreview.net/forum?id=H1edEyBKDS} {Plug and play language models: A simple approach to controlled text generation}.
\newblock In \emph{International Conference on Learning Representations}.

\bibitem[{Deng et~al.(2023)Deng, Zhao, Tang, Gerstein, and Cohan}]{deng2023investigating}
Chunyuan Deng, Yilun Zhao, Xiangru Tang, Mark Gerstein, and Arman Cohan. 2023.
\newblock Investigating data contamination in modern benchmarks for large language models.
\newblock \emph{arXiv preprint arXiv:2311.09783}.

\bibitem[{Ding et~al.(2023{\natexlab{a}})Ding, Chen, Xu, Hu, Qin, Liu, Sun, and Zhou}]{ding_ultrachat_2023}
Ning Ding, Yulin Chen, Bokai Xu, Shengding Hu, Yujia Qin, Zhiyuan Liu, Maosong Sun, and Bowen Zhou. 2023{\natexlab{a}}.
\newblock \href {https://github.com/thunlp/ultrachat} {{UltraChat}: {A} {Large}-scale {Auto}-generated {Multi}-round {Dialogue} {Data}}.
\newblock Published: GitHub Repository.

\bibitem[{Ding et~al.(2023{\natexlab{b}})Ding, Chen, Xu, Qin, Zheng, Hu, Liu, Sun, and Zhou}]{ding_enhancing_2023}
Ning Ding, Yulin Chen, Bokai Xu, Yujia Qin, Zhi Zheng, Shengding Hu, Zhiyuan Liu, Maosong Sun, and Bowen Zhou. 2023{\natexlab{b}}.
\newblock Enhancing {Chat} {Language} {Models} by {Scaling} {High}-quality {Instructional} {Conversations}.
\newblock \emph{arXiv preprint arXiv:2305.14233}.

\bibitem[{Dubois et~al.(2024)Dubois, Galambosi, Liang, and Hashimoto}]{alpaca_eval_length}
Yann Dubois, Balazs Galambosi, Percy Liang, and Tatsunori~B. Hashimoto. 2024.
\newblock Length-corrected alpacaeval: A simple debiasing of automatic evaluators.
\newblock \url{https://github.com/tatsu-lab/alpaca_eval}.

\bibitem[{Gao(2023)}]{gao_tianyu_blog_2023}
Tianyu Gao. 2023.
\newblock \href {https://gaotianyu.xyz/blog/2023/11/30/instruction-tuning} {Teach llamas to talk: Recent progress in instruction tuning}.

\bibitem[{Geng et~al.(2023)Geng, Gudibande, Liu, Wallace, Abbeel, Levine, and Song}]{geng_koala_2023}
Xinyang Geng, Arnav Gudibande, Hao Liu, Eric Wallace, Pieter Abbeel, Sergey Levine, and Dawn Song. 2023.
\newblock Koala: {A} dialogue model for academic research.
\newblock \emph{Blog post, April}, 1.

\bibitem[{Hendrycks et~al.(2021)Hendrycks, Burns, Basart, Critch, Li, Song, and Steinhardt}]{hendrycks_aligning_2021}
Dan Hendrycks, Collin Burns, Steven Basart, Andrew Critch, Jerry Li, Dawn Song, and Jacob Steinhardt. 2021.
\newblock Aligning {AI} {With} {Shared} {Human} {Values}.
\newblock In \emph{{ICLR}}.

\bibitem[{Ivison et~al.(2023)Ivison, Wang, Pyatkin, Lambert, Peters, Dasigi, Jang, Wadden, Smith, Beltagy, and Hajishirzi}]{tulu2}
Hamish Ivison, Yizhong Wang, Valentina Pyatkin, Nathan Lambert, Matthew Peters, Pradeep Dasigi, Joel Jang, David Wadden, Noah~A. Smith, Iz~Beltagy, and Hannaneh Hajishirzi. 2023.
\newblock \href {http://arxiv.org/abs/2311.10702} {Camels in a changing climate: Enhancing lm adaptation with tulu 2}.

\bibitem[{Jiang et~al.(2023{\natexlab{a}})Jiang, Sablayrolles, Mensch, Bamford, Chaplot, Casas, Bressand, Lengyel, Lample, Saulnier, and {others}}]{jiang_mistral_2023}
Albert~Q Jiang, Alexandre Sablayrolles, Arthur Mensch, Chris Bamford, Devendra~Singh Chaplot, Diego de~las Casas, Florian Bressand, Gianna Lengyel, Guillaume Lample, Lucile Saulnier, and {others}. 2023{\natexlab{a}}.
\newblock Mistral {7B}.
\newblock \emph{arXiv preprint arXiv:2310.06825}.

\bibitem[{Jiang et~al.(2023{\natexlab{b}})Jiang, Wang, Zeng, Zhong, Li, Mi, Shang, Jiang, Liu, and Wang}]{jiang_followbench_2023}
Yuxin Jiang, Yufei Wang, Xingshan Zeng, Wanjun Zhong, Liangyou Li, Fei Mi, Lifeng Shang, Xin Jiang, Qun Liu, and Wei Wang. 2023{\natexlab{b}}.
\newblock \href {http://arxiv.org/abs/2310.20410} {{FollowBench}: {A} {Multi}-level {Fine}-grained {Constraints} {Following} {Benchmark} for {Large} {Language} {Models}}.
\newblock \emph{arXiv preprint arXiv:2310.20410}.

\bibitem[{Kopf et~al.(2023)Kopf, Kilcher, Rutte, Anagnostidis, Tam, Stevens, Barhoum, Duc, Stanley, Nagyfi, Shahul, Suri, Glushkov, Dantuluri, Maguire, Schuhmann, Nguyen, and Mattick}]{kopf_openassistant_2023}
Andreas Kopf, Yannic Kilcher, Dimitri~von Rutte, Sotiris Anagnostidis, Zhi~Rui Tam, Keith Stevens, Abdullah Barhoum, Nguyen~Minh Duc, Oliver Stanley, Rich'ard Nagyfi, E.~S. Shahul, Sameer Suri, David Glushkov, Arnav Dantuluri, Andrew Maguire, Christoph Schuhmann, Huu Nguyen, and Alexander Mattick. 2023.
\newblock {OpenAssistant} {Conversations} - {Democratizing} {Large} {Language} {Model} {Alignment}.
\newblock \emph{ArXiv}, abs/2304.07327.

\bibitem[{Li et~al.(2023)Li, Zhang, Dubois, Taori, Gulrajani, Guestrin, Liang, and Hashimoto}]{li_alpacaeval_2023}
Xuechen Li, Tianyi Zhang, Yann Dubois, Rohan Taori, Ishaan Gulrajani, Carlos Guestrin, Percy Liang, and Tatsunori~B. Hashimoto. 2023.
\newblock \href {https://github.com/tatsu-lab/alpaca_eval} {{AlpacaEval}: {An} {Automatic} {Evaluator} of {Instruction}-following {Models}}.
\newblock Publication Title: GitHub repository.

\bibitem[{Lightman et~al.(2023)Lightman, Kosaraju, Burda, Edwards, Baker, Lee, Leike, Schulman, Sutskever, and Cobbe}]{lightman2023let}
Hunter Lightman, Vineet Kosaraju, Yura Burda, Harri Edwards, Bowen Baker, Teddy Lee, Jan Leike, John Schulman, Ilya Sutskever, and Karl Cobbe. 2023.
\newblock Let's verify step by step.
\newblock \emph{arXiv preprint arXiv:2305.20050}.

\bibitem[{Lin et~al.(2022)Lin, Hilton, and Evans}]{lin_truthfulqa_2022}
Stephanie Lin, Jacob Hilton, and Owain Evans. 2022.
\newblock {TruthfulQA}: {Measuring} {How} {Models} {Mimic} {Human} {Falsehoods}.
\newblock In \emph{Proceedings of the 60th {Annual} {Meeting} of the {Association} for {Computational} {Linguistics} ({Volume} 1: {Long} {Papers})}, pages 3214--3252.

\bibitem[{Liu et~al.(2024)Liu, Zeng, He, Jiang, and He}]{liu2023makes}
Wei Liu, Weihao Zeng, Keqing He, Yong Jiang, and Junxian He. 2024.
\newblock \href {https://arxiv.org/abs/2312.15685} {What makes good data for alignment? a comprehensive study of automatic data selection in instruction tuning}.
\newblock In \emph{The Twelfth International Conference on Learning Representations (ICLR)}.

\bibitem[{Longpre et~al.(2023)Longpre, Hou, Vu, Webson, Chung, Tay, Zhou, Le, Zoph, Wei, and {others}}]{longpre_flan_2023}
Shayne Longpre, Le~Hou, Tu~Vu, Albert Webson, Hyung~Won Chung, Yi~Tay, Denny Zhou, Quoc~V Le, Barret Zoph, Jason Wei, and {others}. 2023.
\newblock The flan collection: {Designing} data and methods for effective instruction tuning.
\newblock \emph{arXiv preprint arXiv:2301.13688}.

\bibitem[{Lou et~al.(2024)Lou, Zhang, Xie, Sun, Ahn, Xu, su, and Yin}]{lou_muffin_2023}
Renze Lou, Kai Zhang, Jian Xie, Yuxuan Sun, Janice Ahn, Hanzi Xu, Yu~su, and Wenpeng Yin. 2024.
\newblock \href {https://openreview.net/forum?id=1vrS1zwekw} {{MUFFIN}: Curating multi-faceted instructions for improving instruction following}.
\newblock In \emph{The Twelfth International Conference on Learning Representations (ICLR)}.

\bibitem[{Mishra et~al.(2022)Mishra, Khashabi, Baral, and Hajishirzi}]{mishra_cross-task_2022}
Swaroop Mishra, Daniel Khashabi, Chitta Baral, and Hannaneh Hajishirzi. 2022.
\newblock \href {https://doi.org/10.18653/v1/2022.acl-long.244} {Cross-{Task} {Generalization} via {Natural} {Language} {Crowdsourcing} {Instructions}}.
\newblock In \emph{Proceedings of the 60th {Annual} {Meeting} of the {Association} for {Computational} {Linguistics} ({Volume} 1: {Long} {Papers})}, pages 3470--3487, Dublin, Ireland. Association for Computational Linguistics.

\bibitem[{OpenAI(2022)}]{openai_chatgpt_2022}
OpenAI. 2022.
\newblock Chatgpt: {Optimizing} language models for dialogue.
\newblock \emph{OpenAI}.

\bibitem[{{OpenAI}(2023)}]{openai_gpt-4_2023}
{OpenAI}. 2023.
\newblock {GPT}-4 {Technical} {Report}.
\newblock \emph{CoRR}, abs/2303.08774.
\newblock ArXiv: 2303.08774.

\bibitem[{Ouyang et~al.(2022)Ouyang, Wu, Jiang, Almeida, Wainwright, Mishkin, Zhang, Agarwal, Slama, Ray, and {others}}]{ouyang_training_2022}
Long Ouyang, Jeffrey Wu, Xu~Jiang, Diogo Almeida, Carroll Wainwright, Pamela Mishkin, Chong Zhang, Sandhini Agarwal, Katarina Slama, Alex Ray, and {others}. 2022.
\newblock Training language models to follow instructions with human feedback.
\newblock \emph{Advances in Neural Information Processing Systems}, 35:27730--27744.

\bibitem[{Ouyang et~al.(2023)Ouyang, Wang, Liu, Zhong, Jiao, Iter, Pryzant, Zhu, Ji, and Han}]{ouyang-etal-2023-shifted}
Siru Ouyang, Shuohang Wang, Yang Liu, Ming Zhong, Yizhu Jiao, Dan Iter, Reid Pryzant, Chenguang Zhu, Heng Ji, and Jiawei Han. 2023.
\newblock \href {https://doi.org/10.18653/v1/2023.emnlp-main.146} {The shifted and the overlooked: A task-oriented investigation of user-{GPT} interactions}.
\newblock In \emph{Proceedings of the 2023 Conference on Empirical Methods in Natural Language Processing}, pages 2375--2393, Singapore. Association for Computational Linguistics.

\bibitem[{Peng et~al.(2023)Peng, Li, He, Galley, and Gao}]{peng_instruction_2023}
Baolin Peng, Chunyuan Li, Pengcheng He, Michel Galley, and Jianfeng Gao. 2023.
\newblock Instruction {Tuning} with {GPT}-4.
\newblock \emph{arXiv preprint arXiv:2304.03277}.

\bibitem[{Qin et~al.(2024)Qin, Song, Hu, Yao, Cho, Wang, Wu, Liu, Liu, and Yu}]{qin2024infobench}
Yiwei Qin, Kaiqiang Song, Yebowen Hu, Wenlin Yao, Sangwoo Cho, Xiaoyang Wang, Xuansheng Wu, Fei Liu, Pengfei Liu, and Dong Yu. 2024.
\newblock \href {http://arxiv.org/abs/2401.03601} {Infobench: Evaluating instruction following ability in large language models}.
\newblock \emph{arXiv preprint arXiv:2401.03601}.

\bibitem[{Rafailov et~al.(2023)Rafailov, Sharma, Mitchell, Manning, Ermon, and Finn}]{rafailov_direct_2023}
Rafael Rafailov, Archit Sharma, Eric Mitchell, Christopher~D Manning, Stefano Ermon, and Chelsea Finn. 2023.
\newblock \href {https://openreview.net/forum?id=HPuSIXJaa9} {Direct preference optimization: Your language model is secretly a reward model}.
\newblock In \emph{Thirty-seventh Conference on Neural Information Processing Systems}.

\bibitem[{Sanh et~al.(2021)Sanh, Webson, Raffel, Bach, Sutawika, Alyafeai, Chaffin, Stiegler, Scao, Raja, and {others}}]{sanh_multitask_2021}
Victor Sanh, Albert Webson, Colin Raffel, Stephen~H Bach, Lintang Sutawika, Zaid Alyafeai, Antoine Chaffin, Arnaud Stiegler, Teven~Le Scao, Arun Raja, and {others}. 2021.
\newblock Multitask prompted training enables zero-shot task generalization.
\newblock \emph{arXiv preprint arXiv:2110.08207}.

\bibitem[{Sun et~al.(2023)Sun, Tian, Zhou, Xu, Hu, Gupta, Wieting, Peng, and Ma}]{sun_evaluating_2023}
Jiao Sun, Yufei Tian, Wangchunshu Zhou, Nan Xu, Qian Hu, Rahul Gupta, John Wieting, Nanyun Peng, and Xuezhe Ma. 2023.
\newblock \href {https://doi.org/10.18653/v1/2023.emnlp-main.190} {Evaluating {Large} {Language} {Models} on {Controlled} {Generation} {Tasks}}.
\newblock In \emph{Proceedings of the 2023 {Conference} on {Empirical} {Methods} in {Natural} {Language} {Processing}}, pages 3155--3168, Singapore. Association for Computational Linguistics.

\bibitem[{Taori et~al.(2023)Taori, Gulrajani, Zhang, Dubois, Li, Guestrin, Liang, and Hashimoto}]{taori_stanford_2023}
Rohan Taori, Ishaan Gulrajani, Tianyi Zhang, Yann Dubois, Xuechen Li, Carlos Guestrin, Percy Liang, and Tatsunori~B. Hashimoto. 2023.
\newblock \href {https://github.com/tatsu-lab/stanford_alpaca} {Stanford {Alpaca}: {An} {Instruction}-following {LLaMA} model}.
\newblock Published: GitHub repository.

\bibitem[{Touvron et~al.(2023)Touvron, Martin, Stone, Albert, Almahairi, Babaei, Bashlykov, Batra, Bhargava, Bhosale, and {others}}]{touvron_llama_2023-1}
Hugo Touvron, Louis Martin, Kevin Stone, Peter Albert, Amjad Almahairi, Yasmine Babaei, Nikolay Bashlykov, Soumya Batra, Prajjwal Bhargava, Shruti Bhosale, and {others}. 2023.
\newblock Llama 2: {Open} foundation and fine-tuned chat models.
\newblock \emph{arXiv preprint arXiv:2307.09288}.

\bibitem[{Tunstall et~al.(2023{\natexlab{a}})Tunstall, Beeching, Lambert, Rajani, Huang, Rasul, Rush, and Wolf}]{alignment_handbook2023}
Lewis Tunstall, Edward Beeching, Nathan Lambert, Nazneen Rajani, Shengyi Huang, Kashif Rasul, Alexander~M. Rush, and Thomas Wolf. 2023{\natexlab{a}}.
\newblock The alignment handbook.
\newblock \url{https://github.com/huggingface/alignment-handbook}.

\bibitem[{Tunstall et~al.(2023{\natexlab{b}})Tunstall, Beeching, Lambert, Rajani, Rasul, Belkada, Huang, von Werra, Fourrier, Habib, and {others}}]{tunstall_zephyr_2023}
Lewis Tunstall, Edward Beeching, Nathan Lambert, Nazneen Rajani, Kashif Rasul, Younes Belkada, Shengyi Huang, Leandro von Werra, Clémentine Fourrier, Nathan Habib, and {others}. 2023{\natexlab{b}}.
\newblock Zephyr: {Direct} {Distillation} of {LM} {Alignment}.
\newblock \emph{arXiv preprint arXiv:2310.16944}.

\bibitem[{Wang et~al.(2023)Wang, Cheng, Zhan, Li, Song, and Liu}]{wang_openchat_2023}
Guan Wang, Sijie Cheng, Xianyuan Zhan, Xiangang Li, Sen Song, and Yang Liu. 2023.
\newblock {OpenChat}: {Advancing} {Open}-source {Language} {Models} with {Mixed}-{Quality} {Data}.
\newblock \emph{arXiv preprint arXiv:2309.11235}.

\bibitem[{Wang et~al.(2022)Wang, Kordi, Mishra, Liu, Smith, Khashabi, and Hajishirzi}]{wang_self-instruct_2022}
Yizhong Wang, Yeganeh Kordi, Swaroop Mishra, Alisa Liu, Noah~A Smith, Daniel Khashabi, and Hannaneh Hajishirzi. 2022.
\newblock Self-{Instruct}: {Aligning} {Language} {Model} with {Self} {Generated} {Instructions}.
\newblock \emph{arXiv preprint arXiv:2212.10560}.

\bibitem[{Wei et~al.(2021)Wei, Bosma, Zhao, Guu, Yu, Lester, Du, Dai, and Le}]{wei_finetuned_2021}
Jason Wei, Maarten Bosma, Vincent~Y Zhao, Kelvin Guu, Adams~Wei Yu, Brian Lester, Nan Du, Andrew~M Dai, and Quoc~V Le. 2021.
\newblock Finetuned language models are zero-shot learners.
\newblock \emph{arXiv preprint arXiv:2109.01652}.

\bibitem[{Wei et~al.(2023)Wei, Zhao, Zhang, Zhu, Wang, Yang, Li, Cheng, L{\"u}, Hu et~al.}]{wei2023skywork}
Tianwen Wei, Liang Zhao, Lichang Zhang, Bo~Zhu, Lijie Wang, Haihua Yang, Biye Li, Cheng Cheng, Weiwei L{\"u}, Rui Hu, et~al. 2023.
\newblock Skywork: A more open bilingual foundation model.
\newblock \emph{arXiv preprint arXiv:2310.19341}.

\bibitem[{Xu et~al.(2024)Xu, Sun, Zheng, Geng, Zhao, Feng, Tao, and Jiang}]{xu_wizardlm_2023}
Can Xu, Qingfeng Sun, Kai Zheng, Xiubo Geng, Pu~Zhao, Jiazhan Feng, Chongyang Tao, and Daxin Jiang. 2024.
\newblock Wizardlm: {Empowering} large language models to follow complex instructions.
\newblock In \emph{The Twelfth International Conference on Learning Representations (ICLR)}.

\bibitem[{Yang et~al.(2023)Yang, Chiang, Zheng, Gonzalez, and Stoica}]{yang2023rethinking}
Shuo Yang, Wei-Lin Chiang, Lianmin Zheng, Joseph~E. Gonzalez, and Ion Stoica. 2023.
\newblock \href {http://arxiv.org/abs/2311.04850} {Rethinking benchmark and contamination for language models with rephrased samples}.
\newblock \emph{arXiv preprint arXiv:2311.04850}.

\bibitem[{Zellers et~al.(2019)Zellers, Holtzman, Bisk, Farhadi, and Choi}]{zellers_hellaswag_2019}
Rowan Zellers, Ari Holtzman, Yonatan Bisk, Ali Farhadi, and Yejin Choi. 2019.
\newblock {HellaSwag}: {Can} a {Machine} {Really} {Finish} {Your} {Sentence}?
\newblock In \emph{The Annual Meeting of the Association for Computational Linguistics (ACL)}.

\bibitem[{Zhang et~al.(2023{\natexlab{a}})Zhang, Song, Li, Zhou, and Song}]{CTGSurvey2023}
Hanqing Zhang, Haolin Song, Shaoyu Li, Ming Zhou, and Dawei Song. 2023{\natexlab{a}}.
\newblock A survey of controllable text generation using transformer-based pre-trained language models.
\newblock \emph{ACM Computing Surveys}.

\bibitem[{Zhang et~al.(2023{\natexlab{b}})Zhang, Dong, Li, Zhang, Sun, Wang, Li, Hu, Zhang, Wu, and Wang}]{instruction_tuning_survey}
Shengyu Zhang, Linfeng Dong, Xiaoya Li, Sen Zhang, Xiaofei Sun, Shuhe Wang, Jiwei Li, Runyi Hu, Tianwei Zhang, Fei Wu, and Guoyin Wang. 2023{\natexlab{b}}.
\newblock \href {http://arxiv.org/abs/2308.10792} {Instruction tuning for large language models: {A} survey}.
\newblock \emph{arXiv preprint arXiv:2308.10792}.

\bibitem[{Zhang et~al.(2020)Zhang, Wang, Li, Gan, Brockett, and Dolan}]{zhang_pointer_2020}
Yizhe Zhang, Guoyin Wang, Chunyuan Li, Zhe Gan, Chris Brockett, and William~B Dolan. 2020.
\newblock Pointer: {Constrained} {Text} {Generation} via {Insertion}-based {Generative} {Pre}-training.
\newblock In \emph{Proceedings of the 2020 {Conference} on {Empirical} {Methods} in {Natural} {Language} {Processing} ({EMNLP})}, pages 8649--8670.

\bibitem[{Zhao et~al.(2024)Zhao, Andriushchenko, Croce, and Flammarion}]{zhao2024long}
Hao Zhao, Maksym Andriushchenko, Francesco Croce, and Nicolas Flammarion. 2024.
\newblock \href {http://arxiv.org/abs/2402.04833} {Long is more for alignment: A simple but tough-to-beat baseline for instruction fine-tuning}.

\bibitem[{Zheng et~al.(2023)Zheng, Chiang, Sheng, Zhuang, Wu, Zhuang, Lin, Li, Li, Xing, Zhang, Gonzalez, and Stoica}]{zheng_judging_2023}
Lianmin Zheng, Wei-Lin Chiang, Ying Sheng, Siyuan Zhuang, Zhanghao Wu, Yonghao Zhuang, Zi~Lin, Zhuohan Li, Dacheng Li, Eric~P. Xing, Hao Zhang, Joseph~E. Gonzalez, and Ion Stoica. 2023.
\newblock Judging {LLM}-as-a-judge with {MT}-{Bench} and {Chatbot} {Arena}.
\newblock In \emph{{NeurIPS} {Datasets} and {Benchmarks} {Track}}.
\newblock \_eprint: 2306.05685.

\bibitem[{Zhou et~al.(2023{\natexlab{a}})Zhou, Liu, Xu, Iyer, Sun, Mao, Ma, Efrat, Yu, Yu, and {others}}]{zhou_lima_2023}
Chunting Zhou, Pengfei Liu, Puxin Xu, Srini Iyer, Jiao Sun, Yuning Mao, Xuezhe Ma, Avia Efrat, Ping Yu, Lili Yu, and {others}. 2023{\natexlab{a}}.
\newblock Lima: {Less} is more for alignment.
\newblock \emph{arXiv preprint arXiv:2305.11206}.

\bibitem[{Zhou et~al.(2023{\natexlab{b}})Zhou, Lu, Mishra, Brahma, Basu, Luan, Zhou, and Hou}]{zhou_instruction-following_2023}
Jeffrey Zhou, Tianjian Lu, Swaroop Mishra, Siddhartha Brahma, Sujoy Basu, Yi~Luan, Denny Zhou, and Le~Hou. 2023{\natexlab{b}}.
\newblock Instruction-{Following} {Evaluation} for {Large} {Language} {Models}.
\newblock \emph{arXiv preprint arXiv:2311.07911}.

\end{thebibliography}
\clearpage
\appendix

\label{sec:appendix}

\section{Details of \name Dataset}
\label{app:dataset}

\subsection{Dataset Filtering Cases}
\begin{table*}[th]
    \setlength{\belowcaptionskip}{-0.4cm}
    \centering
    \small
    \begin{tabular}{l|cp{11cm}}
        \toprule
        \multicolumn{3}{c}{\textbf{First Filter} } \\
        \midrule
        \multirow{6}{*}{Case 1}
         & \textbf{Query}: & How does the noise cancellation technology in the Bose QuietComfort 35 II compare to other models? \\
         & \textbf{Filter Response}: & No, because providing a comparison \textbf{requires up-to-date} and specific information about other models' noise cancellation technology, which may have evolved since my last update, and without knowing which specific models you're referring to, a detailed comparison cannot be made. \\
        \midrule
        \multirow{3}{*}{Case 2}
        & \textbf{Query}: & What is the total amount of payroll if 75\% is paid in checks? \\
        & \textbf{Filter Response}: & No, because the total amount of payroll is not specified, and \textbf{without knowing the total or the actual amount} paid in checks, it is impossible to calculate the total payroll. \\
        \midrule
        
        \multicolumn{3}{c}{\textbf{Final Filter} } \\
        \midrule
        \multirow{5}{*}{Case 1}
        & \textbf{Instruction}: & What technical manuals offer guidance on using paragraphs versus numbered steps for novice users? \\
        & \textbf{Filter Response}: & No, because the question is asking for specific technical manuals, which could be numerous and varied across different fields and industries. \textbf{Without additional context} about the field or type of technology, it is not possible to provide a specific answer. \\
        \midrule
        \multirow{8}{*}{Case 2}
        & \textbf{Instruction}: & Considering a sensor network application, what are the necessary ESP32 module technical specifications, antenna design, and ESP-NOW protocol features, including pairing and encryption? Limit to 7 items. \\
        & \textbf{Filter Response}: & No, because the question is asking for a list of specifications and features \textbf{that are too broad and would require more than 7 items} to be fully addressed. Additionally, the question combines different aspects such as technical specifications, antenna design, and protocol features, which are complex and cannot be effectively condensed into a limited list without additional context or prioritization. \\
        \bottomrule
    \end{tabular}
    \caption{The cases that have been filtered after the two-stage filtering, we list the original queries and why they are filtered. The filter response is generated by GPT-4 Turbo. The reasons that the query/instruction cannot be answered are highlighted.}
    \label{tab:filter_case_study}
\end{table*}
Table \ref{tab:filter_case_study} demonstrates cases filtered through the two-stage filtering process, and it includes explanations from GPT-4 Turbo on why certain cases were removed.

\section{Details of Experiments}
\label{app:exp}
\subsection{Details of Training}
\label{app:training}
For model training, we utilize the widely-used repository `The Alignment Handbook' \cite{alignment_handbook2023}, released by  HuggingFaceH4, to train Mistral-7B based models, and use FastChat \cite{zheng_judging_2023} to train LLaMA-2-13B based models. The Mistral-7B based experiments in the paper are done using 8 NVIDIA A100 80GB GPUs. We adopt DeepSpeed ZeRO stage 2 for SFT and DPO training. The SFT models are trained with 4 epochs. We set the learning rate to $2.0e-05$, with per device batch size 16 and gradient accumulation steps 4. The warm-up ratio is set to 0.1. While LLaMA-2-13B based experiments are done using 16 NVIDIA A100 80GB GPUs, trained with 3 epochs, per device batch size 2, gradient accumulation steps 4, and warm-up ratio is set to 0.03. For DPO training, the beta is set to 0.01 and trained with 1 epoch. The learning rate is set to $5.0e-07$, with per device batch size 4 and gradient accumulation steps 4. We apply cosine learning rate scheduler and gradient checkpointing in the experiments and the max sequence length is set to 2048. 

\subsection{Details of Baselines}
The baseline methods in the experiments include: 
\begin{itemize}
    \item \textbf{Qwen-72B-Chat} \cite{qwen}, developed by Alibaba Cloud, is among the best open-source LLMs, which is pretrained on 3T tokens. However, the dataset adopted for alignment of the model is undisclosed.
    \item \textbf{LLaMA-2-70B-Chat} \cite{touvron_llama_2023-1} is a powerful open-source model released by Meta. It a fine-tuned version of LLaMA-2-70B that is optimized using SFT and RLHF for dialogue use cases. The datasets for training this model is closed-source.
    \item \textbf{Vicuna-13B-v1.5} \cite{chiang_vicuna_2023} is a powerful LLaMA-2-13B based model which trained on 125k ShareGPT data, but the training data will not release.
    \item \textbf{Zephyr-7B-beta} \cite{tunstall_zephyr_2023} is a state-of-the-art LLM based on Mistral-7B. Zephyr-7B-beta has been fine-tuned on UltraChat \cite{ding_enhancing_2023} instruction tuning dataset and optimized with DPO on the UltraFeedback \cite{cui_ultrafeedback_2023} dataset.
    \item \textbf{Deita-7B-1.0-SFT} is another state-of-the-art LLM based on Mistral-7B. Deita proposes a data selection strategy and curate a 6K high-quality subset from ShareGPT, UltraChat, and WizardLM Evol-Instruct \cite{xu_wizardlm_2023} datasets. So the general SFT data is better than ShareGPT used in Conifer. However, Conifer can be further combined with data selection methods like Deita to achieve better overall performances. \textbf{Deita-7B-V1.0} \cite{liu2023makes} is further aligned using DPO on the UltraFeedback dataset.
    \item \textbf{LLaMA-2-13B-ShareGPT} and \textbf{Mistral-7B-ShareGPT} are models we train with the publicly available 53k ShareGPT instruction tuning dataset, using LLaMA-2-13B and Mistral-7B as base models respectively. We further apply DPO training to get \textbf{Mistral-7B-ShareGPT-DPO} under the
    same UltraFeedback dataset and experimental conditions as \texttt{Conifer} for fair comparisons.
    \item \textbf{Mistral-7B-Evol-Instruct} \cite{xu_wizardlm_2023}: We have trained this model on the publicly available WizardLM-Evol-Instruct dataset\footnote{\url{https://huggingface.co/datasets/WizardLM/WizardLM_evol_instruct_V2_196k}}, with 143k sample containing mixture evolved data of Alpaca and ShareGPT. We follow the original paper \cite{xu_wizardlm_2023} to combine 53k ShareGPT dataset with 143k evol-instruct dataset to get final 196k full dataset.
    \item \textbf{Mistral-7B-Muffin}: We trained the model using the Muffin dataset, with a combination with ShareGPT dataset to train Mistral-7B for fair comparison with \texttt{Conifer}. We also try only adopt Muffin dataset and get inferior results. Muffin is an instruction-following dataset aligning with scaling tasks per input. Multi-faceted instructions are curated with the assistance of ChatGPT and GPT-4.
\end{itemize}

\begin{table*}[t]
\centering
\setlength\tabcolsep{3.5pt} 
\begin{tabular}{lccccccc}
\toprule
\multicolumn{1}{c}{Model} & Average & \multicolumn{1}{c}{\begin{tabular}[c]{@{}c@{}}MMLU\\ (10 shot)\end{tabular}} & \multicolumn{1}{c}{\begin{tabular}[c]{@{}c@{}}ARC\\ (25 shot)\end{tabular}} & \multicolumn{1}{c}{\begin{tabular}[c]{@{}c@{}}HellaSwag\\ (10 shot)\end{tabular}} & \multicolumn{1}{c}{\begin{tabular}[c]{@{}c@{}}TruthfulQA\\ (0 shot)\end{tabular}} & \multicolumn{1}{c}{\begin{tabular}[c]{@{}c@{}}WinoGrande\\ (5 shot)\end{tabular}} & \multicolumn{1}{c}{\begin{tabular}[c]{@{}c@{}}GSM8k\\ (5 shot)\end{tabular}}\\
\midrule
ShareGPT     &   59.68           & 59.43                                                                         & 57.34                                                                       & 81.66                                                                             & 52.30    & 71.98   & 35.41                                                   \\
Conifer    & 59.60               & 58.88                                                                        & 58.28                                                                       & 81.80                                                                             & 49.31     & 71.82   & 37.53                                                    \\
\bottomrule
\end{tabular}
\caption{Results on the Open LLM leaderboard.}
\label{tab:open_llm}
\end{table*}

\begin{table*}[t]
\centering
\resizebox{\textwidth}{!}{%
\begin{tabular}{ll|cccc|cccccc}
\toprule
\multirow{2}{*}{Model} & \multirow{2}{*}{Base Model}   & \multicolumn{4}{c|}{IFEval}       & \multicolumn{6}{c}{FollowBench (SSR metric) }                                                                                                                                                                              \\
\cmidrule(lr){3-6} \cmidrule(lr){7-12}
\multicolumn{1}{c}{}      & \multicolumn{1}{c}{}           & \multicolumn{1}{|c}{strict prompt} 
&\multicolumn{1}{c}{loose prompt}
&\multicolumn{1}{c}{strict instruction}
&\multicolumn{1}{c|}{loose instruction}
& \multicolumn{1}{c}{Level 1} & \multicolumn{1}{c}{Level 2} & \multicolumn{1}{c}{Level 3} & \multicolumn{1}{c}{Level 4} & \multicolumn{1}{c}{Level 5} & \multicolumn{1}{c}{Avg}  \\
\midrule
GPT-4$^\dagger$                    & -          & 76.9          & 79.3         & 83.6       & 85.4      & 84.7    & 77.6    & 76.2    & 77.9    & 73.3    & 77.9 \\
GPT-3.5 Turbo$^\dagger$            & -          & -             & -            & -          & -         & 80.3    & 71.2    & 74.2    & 69.6    & 67.1    & 72.5 \\
\midrule
Qwen-72B-Chat$^\dagger$            & Qwen       & -             & 50.8         & -          & -         & 73.8    & 67.5    & 63.2    & 57.6    & 56.0    & 63.6 \\
LLaMA-2-70B-Chat$^\dagger$         & LLaMA-2    & -             & -            & -          & -         & 59.9    & 57.3    & 55.7    & 53.3    & 53.2    & 55.9 \\
\midrule
Vicuna-13B-v1.5          & LLaMA-2    & 43.1          & 46.6         & 53.6       & 58.0      & 69.1    & 64.5    & 58.6    & 52.6    & 52.7    & 59.5 \\
LLaMA-2-13B-ShareGPT     & LLaMA-2    & 42.9          & 47.1         & 53.7       & 57.1      & 59.2    & 55.4    & 57.0    & 48.1    & 49.6    & 53.9 \\
Conifer-13B              & LLaMA-2    & 42.9          & 47.5         & 53.0       & 57.4      & 60.5    & 58.3    & 58.2    & 53.9    & 51.1    & 56.4 \\
\midrule
Deita-7B-v1.0-SFT        & Mistral    & 42.0          & 45.1         & 54.3       & 57.3      & 55.8    & 58.5    & 54.0    & 47.6    & 51.4    & 53.5 \\
Zephyr-7B-beta           & Mistral    & 32.0          & 44.9         & 46.8       & 58.0      & 57.6    & 55.9    & 54.2    & 54.3    & 48.5    & 54.1 \\
Deita-7B-v1.0            & Mistral    & 44.6          & 51.9         & 56.6       & 63.7      & 55.8    & 54.8    & 54.7    & 52.4    & 53.7    & 54.3 \\
Mistral-7B-Muffin        & Mistral    & 32.9          & 34.0         & 44.0       & 45.4      & 40.1    & 39.7    & 37.9    & 35.7    & 36.7    & 38.0 \\
Mistral-7B-Evol-Instruct & Mistral    & 41.4          & 44.0         & 51.3       & 54.4      & 53.2    & 57.0    & 54.7    & 51.2    & 47.4    & 52.7 \\
Mistral-7B-ShareGPT      & Mistral    & 37.5          & 43.4         & 49.3       & 54.9      & 55.7    & 56.6    & 53.6    & 53.4    & 49.7    & 53.8 \\
Conifer-7B               & Mistral    & 45.8          & 50.8         & 57.1       & 62.0      & 53.9    & 57.6    & 53.7    & 54.5    & 49.7    & 53.9 \\
Mistral-7B-ShareGPT-DPO  & Mistral    & 43.8          & 48.2         & 55.8       & 59.7      & 58.4    & 58.1    & 55.8    & 54.9    & 52.3    & 55.9 \\
Conifer-7B-DPO           & Mistral    & 48.1          & 52.3         & 59.1       & 63.3      & 60.3    & 55.7    & 55.7    & 55.9    & 53.3    & 56.2 \\
\bottomrule
\end{tabular}%
}
\caption{Additional results on two instruction following benchmarks: IFEval and FollowBench. $^\dagger$ indicates that the results are directly sourced from the original benchmarks.}
\label{tab:ins_fol_benchmark_appendix}
\end{table*}

\subsection{Results on the Open LLM Leaderboard}
Several traditional studies on instruction tuning \cite{longpre_flan_2023, sanh_multitask_2021, lou_muffin_2023} that aggregate data from a wide array of NLP tasks to train multi-task models, mainly evaluate models on downstream NLP benchmarks. There has been a notable discrepancy shown between these NLP task performances and human preferences \cite{gao_tianyu_blog_2023, zheng_judging_2023, alpaca_eval_length}. Among them, Open LLM leaderboard is a widely used evaluation benchmark. It is important to recognize that leaderboard evaluates abilities that are less correlated with instruction-following ability of chat models \cite{tunstall_zephyr_2023}, or human preferences \cite{alpaca_eval_length}. Although Open LLM leaderboard evaluates capabilities that  are outside the primary focus of this paper, we also following previous work \cite{tunstall_zephyr_2023}, report the results to validate whether fine-tuning has introduced regressions on the model’s reasoning and truthfulness capabilities. 

The Open LLM Leaderboard consists of six tasks, including MMLU \cite{hendrycks_aligning_2021}, HellaSwag \cite{zellers_hellaswag_2019}, ARC \cite{clark_think_2018}, TruthfulQA \cite{lin_truthfulqa_2022}, WinoGrande\cite{ai2:winogrande} and GSM8k \cite{cobbe2021gsm8k}. As is shown in Table \ref{tab:open_llm}, compared to Mistral-7B-ShareGPT baseline (ShareGPT for brevity), incorporating Conifer dataset has shown almost identical average score on the leaderboard. This outcome confirms that the inclusion of the Conifer dataset does not compromise on the model's performance on these benchmarks.

\subsection{Full Results on the IFEval and FollowBench benchmarks}
\label{app:ins_fol_benchmark_appendix}
\begin{figure*}[t] 
	\centering
        \subfigure[LLaMA HSR]{
	    \centering
		\label{fig:fb_hsr_llama}
		\includegraphics[width=0.45\linewidth]{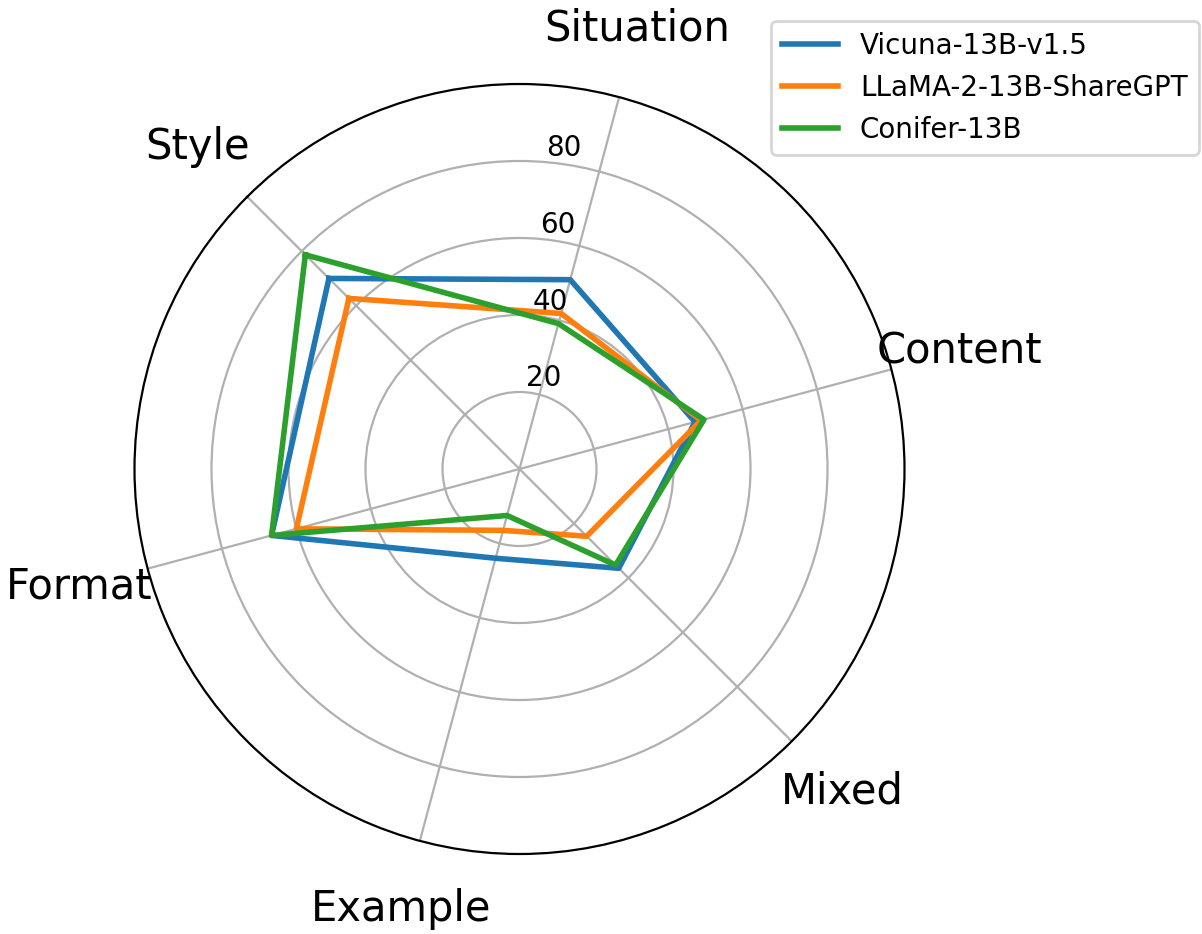}
        }\quad
	\subfigure[LLaMA SSR]{
	    \centering
		\label{fig:fb_ssr_llama}
		\includegraphics[width=0.45\linewidth]{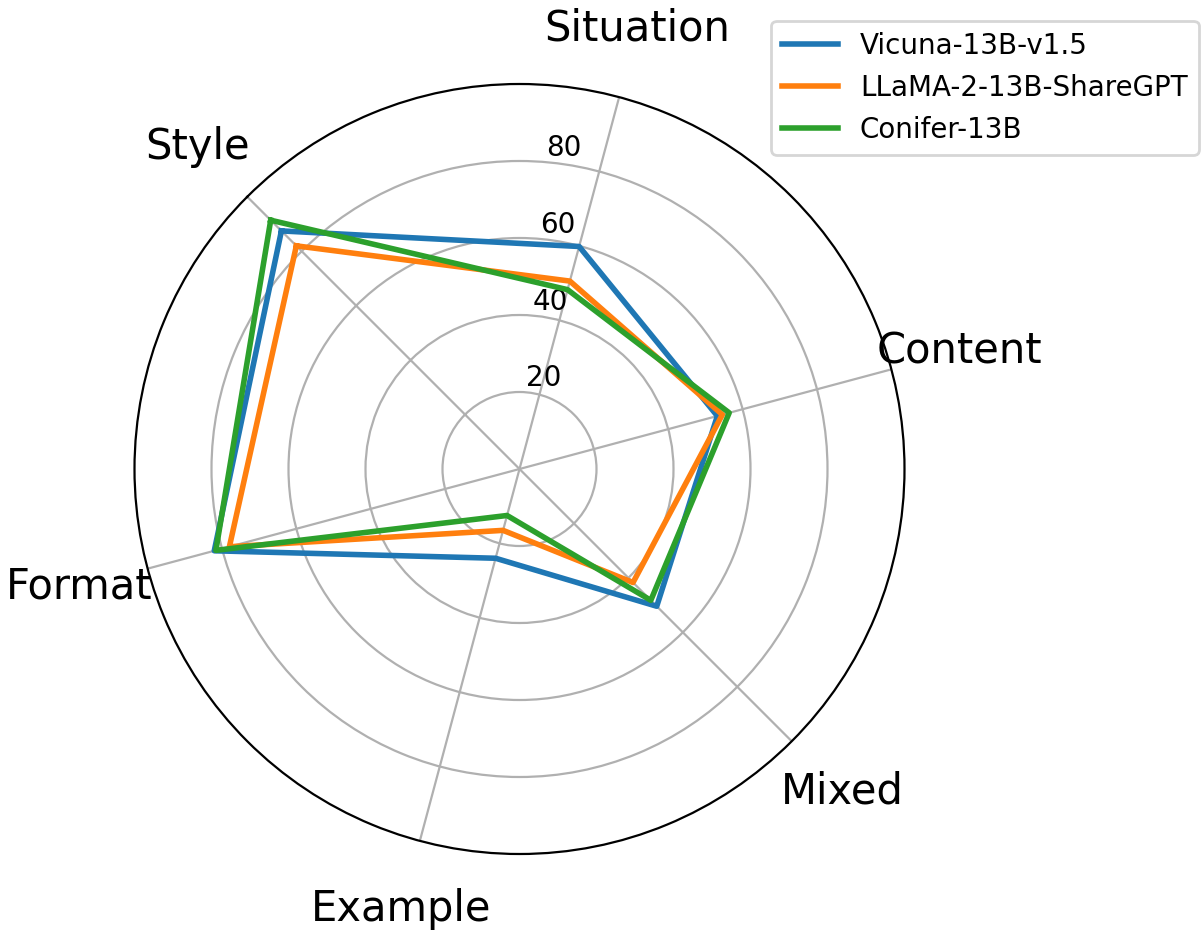}
        }\\
	\subfigure[Mistral HSR]{
	    \centering
		\label{fig:fb_hsr_mistral}
		\includegraphics[width=0.45\linewidth]{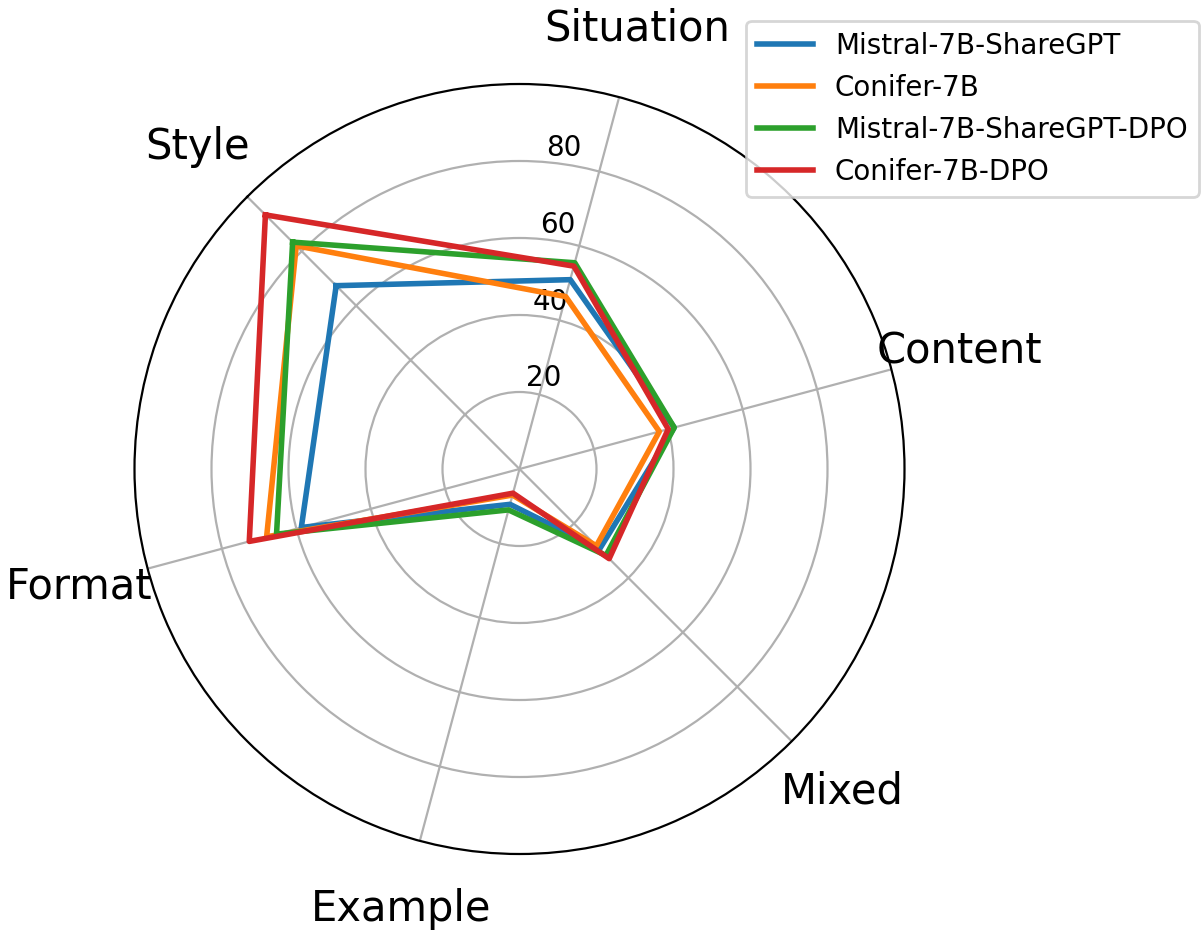}
        }\quad
	\subfigure[Mistral SSR]{
	    \centering
		\label{fig:fb_ssr_mistral}
		\includegraphics[width=0.45\linewidth]{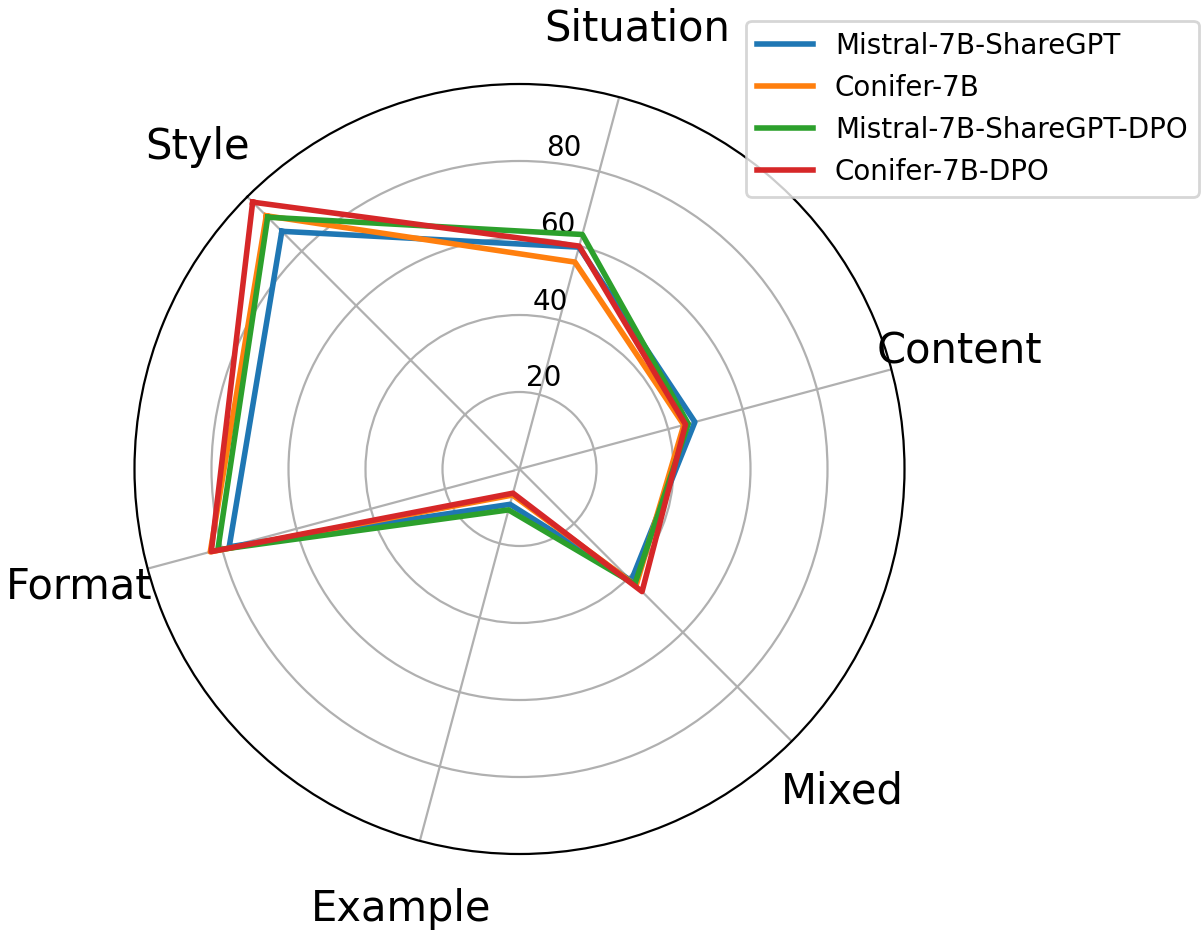}
        }
	\caption{Average HSR (\%) and SSR (\%) results across different levels in diverse constraint categories of our models and baselines.}
	\label{fig:fb_categories}
\end{figure*}

We report IFEval results under the loose prompt setting and FollowBench results under the Hard Satisfaction Rate (HSR) metric in Table \ref{tab:ins_fol_benchmark}. The other metrics, including IFEval under strict prompt, loose instruction and strict instruction settings, and FollowBench under Soft Satisfaction
Rate (SSR) metric are detailed in Table \ref{tab:ins_fol_benchmark_appendix}. We also illustrate the FollowBench scores on the contained six categories as shown in Figure \ref{fig:fb_categories}. These metric results are in agreement with the results shown in Table \ref{tab:ins_fol_benchmark}. 

\end{document}